\documentclass[a4paper,fleqn]{cas-dc}

\usepackage[authoryear]{natbib}
\usepackage{caption}
\usepackage{subcaption}
\def\tsc#1{\csdef{#1}{\textsc{\lowercase{#1}}\xspace}}
\tsc{WGM}
\tsc{QE}
\begin{document}
\let\WriteBookmarks\relax
\def\floatpagepagefraction{1}
\def\textpagefraction{.001}

% Short title
\shorttitle{A Deep-Learning Model for Heterogeneous Precipitation Data Downscaling}    

% Short author
 \shortauthors{A Deep-Learning Model for Heterogeneous Precipitation Data Downscaling} 

% Main title of the paper
\title [mode = title]{Climate Downscaling: A Deep-Learning Based Super-resolution Model of Precipitation Data with Attention Block and Skip Connections}

% Title footnote mark
% eg: \tnotemark[1]
% \tnotemark[<tnote number>] 

% Title footnote 1.
% eg: \tnotetext[1]{Title footnote text}
\tnotetext[t1]{The authors are with the Department of Computer Science and Information Engineering, National Taiwan Normal University, Taipei, Taiwan.} 

% First author
%
% Options: Use if required
\author{Chia-Hao Chiang$^a$}[type=editor,
      style=chinese,
      auid=001,
 ]
 
\author{Zheng-Han Huang$^a$}[type=editor,
      style=chinese,
]

\author{Liwen Liu$^a$}[type=editor,
      style=chinese,
]

\author{Hsin-Chien Liang$^b$}[type=editor,
      style=chinese,
]

\author{Yi-Chi Wang$^b$}[type=editor,
      style=chinese,
]

\author{Wan-Ling Tseng$^c$}[type=editor,
      style=chinese,
]

\author{Chao Wang$^a$}[type=editor,
      style=chinese,
]

\author{Che-Ta Chen$^d$}[type=editor,
      style=chinese,
]

\author{Ko-Chih Wang$^{a,*}$}[type=editor,
      style=chinese,
      auid=000,
      bioid=1,
      orcid=0000-0002-7241-1939,]

\let\printorcid\relax

% Corresponding author indication
% \cormark[]

% Footnote of the first author
% \fnmark[<footnote mark no>]

\ead{kcwang@ntnu.edu.tw}
% \ead{60947072S@ntnu.edu.tw}

% URL of the first author
% \ead[url]{<URL>}

% Credit authorship
% eg: \credit{Conceptualization of this study, Methodology, Software}
% \credit{<Credit authorship details>}

% Address/affiliation
\affiliation{organization={$^a$Computer Science and Information Engineering, National Taiwan Normal University},
            % addressline={}, 
            city={Taipei},
            postcode={116}, 
            % state={},
            country={Taiwan}}

% Research Center for Environmental Changes, Academia Sinica, Taipei, Taiwan
\affiliation{organization={$^b$Research Center for Environmental Changes, Academia Sinica},
            % addressline={}, 
            city={Taipei},
            postcode={116}, 
            % state={},
            country={Taiwan}}

% International Degree Program in Climate Change and Sustainable Development, National Taiwan University
\affiliation{organization={$^c$Degree Program in Climate Change and Sustainable Development, National Taiwan University},
            % addressline={}, 
            city={Taipei},
            postcode={116}, 
            % state={},
            country={Taiwan}}

\affiliation{organization={$^d$Department of Earth Science, National Taiwan University},
            % addressline={}, 
            city={Taipei},
            postcode={116}, 
            % state={},
            country={Taiwan}}

% \author[<aff no>]{<author name>}[<options>]

% Footnote of the second author
% \fnmark[2]

% Email id of the second author
% \ead{}

% URL of the second author
% \ead[url]{}

% Credit authorship
% \credit{}

% Address/affiliation
% \affiliation[<aff no>]{organization={},
%             addressline={}, 
%             city={},
% %          citysep={}, % Uncomment if no comma needed between city and postcode
%             postcode={}, 
%             state={},
%             country={}}

% Corresponding author text
\cortext[1]{Corresponding author}

% Footnote text
% \fntext[1]{}

% For a title note without a number/mark
%\nonumnote{}

% Here goes the abstract
% TODO: Seperate each diagrams
\begin{abstract}
Human activities accelerate consumption of fossil fuels and produce greenhouse gases, resulting in urgent issues today: global warming and the climate change. 
These indirectly cause severe natural disasters, plenty of lives suffering and huge losses of agricultural properties. 
To mitigate impacts on our lands, scientists are developing renewable, reusable, and clean energies and climatologists are trying to predict the extremes. 
Meanwhile, governments are publicizing resource-saving policies for a more eco-friendly society and arousing environment awareness. 
One of the most influencing factors is the precipitation, bringing condensed water vapor onto lands. 
Water resources are the most significant but basic needs in society, not only supporting our livings, but also economics. 
In Taiwan, although the average annual precipitation is up to 2,500 millimeter (mm), the water allocation for each person is lower than the global average due to drastically geographical elevation changes and uneven distribution through the year. 
Thus, it is crucial to track and predict the rainfall to make the most use of it and to prevent the floods. 
However, climate models have limited resolution and require intensive computational power for local-scale use. 
Therefore, we proposed a deep convolutional neural network with skip connections, attention blocks, and auxiliary data concatenation, in order to downscale the low-resolution precipitation data into high-resolution one. 
Eventually, we compare with other climate downscaling methods and show better performance in metrics of  Mean Absolute Error (MAE), Root Mean Square Error (RMSE), Pearson Correlation, structural similarity index (SSIM), and forecast indicators.
\end{abstract}

% Use if graphical abstract is present
%\begin{graphicalabstract}
%\includegraphics{}
%\end{graphicalabstract}

% Research highlights
% \begin{highlights}
% \item 
% \item 
% \item 
% \end{highlights}

% Keywords
% Each keyword is seperated by \sep

\maketitle

% Main text
\section{Introduction}
\label{sec:intro}

%% ------------------------------ The energy issue on Earth -------------------------------------%%
In the past century, human activities have accelerated the consumption of fossil fuels and the emissions of greenhouse gases have drastically increased. 
Accordingly, the global warming has become an inevitable issue and it relates to the climate changes: desertification, severe droughts, wildfires, flooding to name but a few, showing the climate system has become unstable and irregular. 
To study climate changes, such as the precipitation trends in the future, one of the common approaches is to develop simulation models. 
Scientists develop several models to predict climate events to mitigate the impacts of these extremes (website: \cite{ClimateModels}). 
\defcitealias{GCMs}{GCMs}
Among these models, Global Climate Models (\citetalias{GCMs})  are most widely used by climatologists. 
GCMs are used to simulate future climate patterns by considering the interactions among energy systems of lands, oceans, and atmosphere levels via sophisticated physical and mathematical processes. 
Although GCMs can provide a long-term, comprehensive climate trend, the computational cost is relatively high. 
Therefore, scientists often simulate the long-term climate trend with a low-resolution setting to reduce the costs and then feed the GCM results into the climate downscaling methods (e.g., \cite{DownGCM, DownGCM01, DownGCM02, DownGCM03}) to obtain finer results for local-scale uses. 
Another way to provide more accurate results is to adopt Regional Climate Models (RCMs), as proposed by \cite{RCMs}. These models are capable of generating even higher spatial resolution up to tens of kilometers between data points. 
Aside from the computational cost, RCMs are not likely to provide general climate patterns since they are sensitive to the given boundary conditions and regional scale forcings (such as land-sea contrast, orography). 
Some studies show that RCMs perform better in continental regions such as the Great Plains and China, where the forcings are weaker, but perform worse in those with stronger forcings such as New Zealand (e.g., \cite{RCMinCont}).
As a result, for even higher resolution needs (several kilometers), a more accurate, skillful method is required for the climate downscaling.

%% --------------------------------------------- SR ----------------------------------------------%%
Think beyond the climatology scope, recent studies have addressed the super-resolution (SR) techniques to solve the climate downscaling problems (e.g., \cite{DeepSD, deepsdindia, ResDeepD, FSRCNN-ESM, ResLap, YNet}). 
Conceptually, SR is equivalent to climate downscaling. 
SR has been a classic but sophisticated problem in the computer vision field for years. 
It aims to generate high-resolution images from given low-resolution images. 
In the following paragraphs, we will use ``image upscaling'' or ``climate downscaling'' to avoid ambiguity in the process of enlarging the resolution of data.

%% ------------------------------------------ CNN ------------------------------------------------%%
SR models are mainly built with Convolutional Neural Networks (CNNs) which have been widely used in classification, object detection, natural language processing, etc. 
Convolutional layers often act as feature extractors and non-linear mappings (with activation functions) to capture information in different levels. 
Combined with interpolation layers, it is able to reconstruct detailed information from low- to high-resolution images, and it outperforms conventional interpolation methods (e.g., \cite{SRCNN, FSRCNN, VDSR, ESPCN, DRCN, EDSR, LapSRN, SRGAN}).

\begin{figure}[t]
    \centering
        % Topology
        \begin{subfigure}[t]{0.45\columnwidth}
            \includegraphics[width=1\columnwidth]{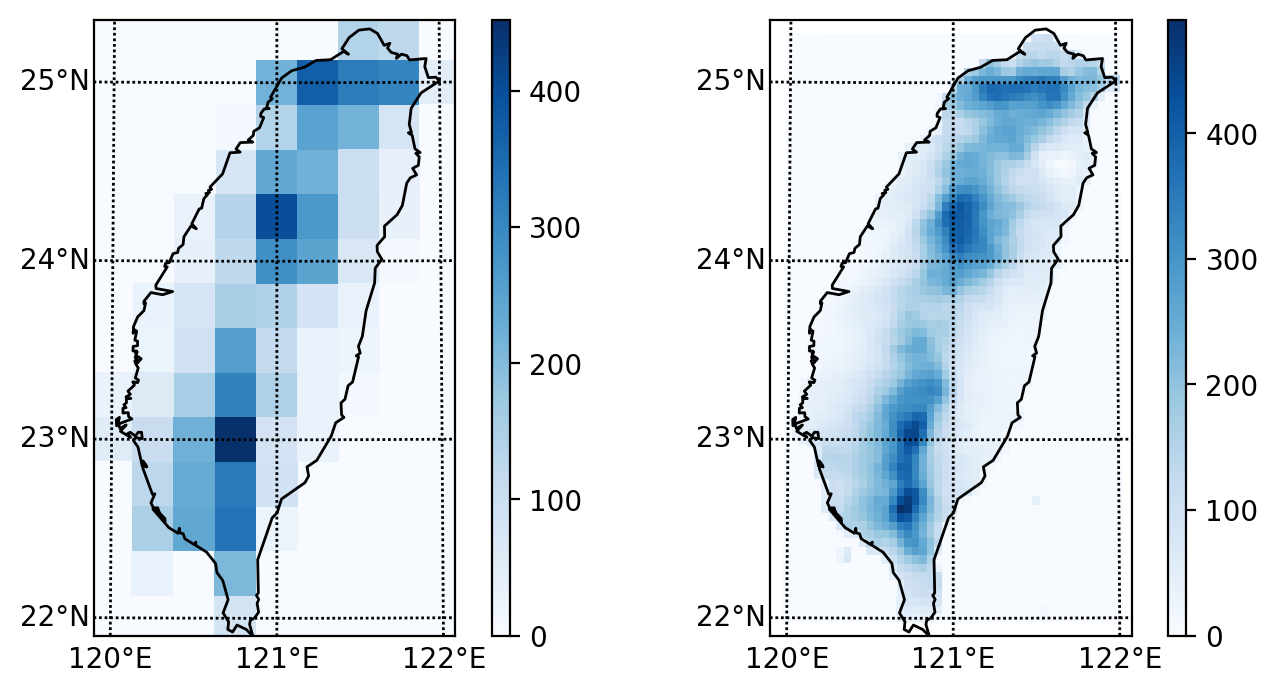}
            \caption{Homogeneity of Dataset}
        \end{subfigure}
        \hfill
        % Divergence
        \begin{subfigure}[t]{0.45\columnwidth}
            \includegraphics[width=1\columnwidth]{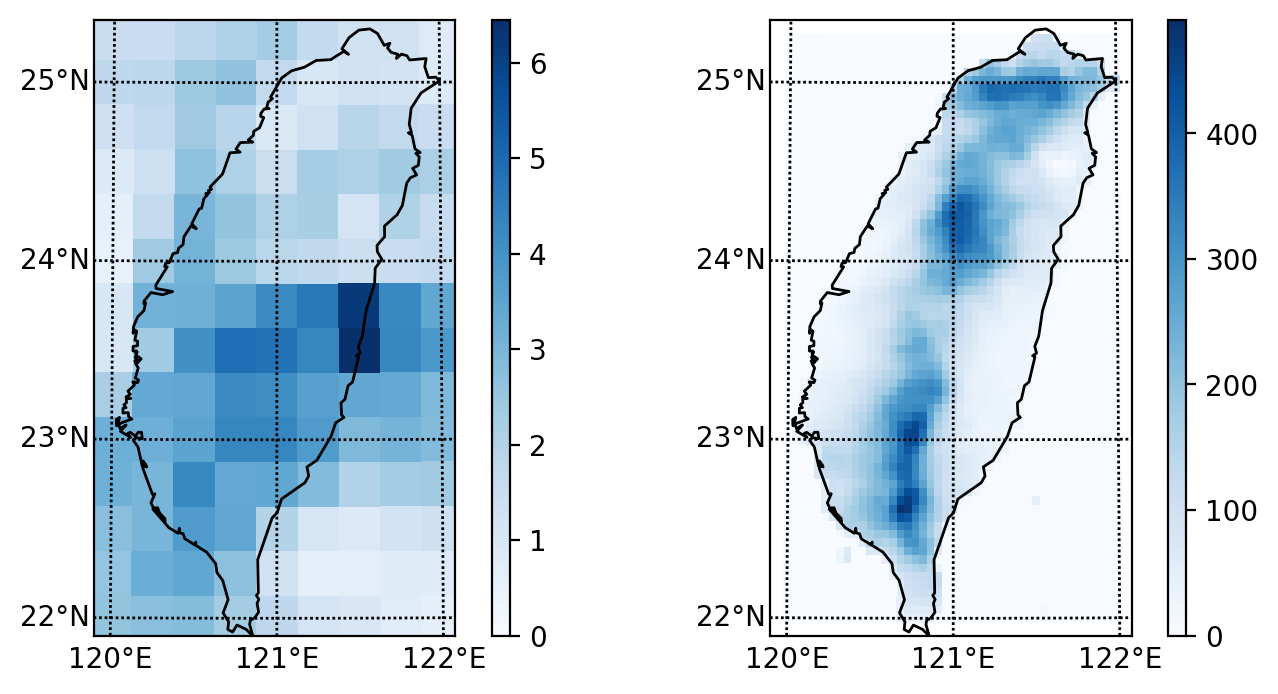}
            \caption{Heterogeneity of Dataset}
        \end{subfigure}
        \hfill
    \caption{Homogeneity and heterogeneity of the datasets. 
            (a) Homogeneous data pair. One can still tell that the low-resolution one (downsampled one, on the left-hand side) is the similar to the high-resolution one (original data, on the right-hand side). 
            The downsampled data is structural similar to the origin one, that there are same patterns at corresponding regions, but only with some lacks of high-frequency information in low-resolution one.
            (b) Heterogeneous data pair. It is hard to tell the structural similarity between them through patterns, since one is from ERA5 reanalysis data (left-hand side), and the other is from observations (right-hand side).
            Neither patterns nor the value ranges are similar.}
    \label{fig:homoheterdataset}
\end{figure}

%% --------------------------- ML/DL based climate downscaling model -----------------------------%%
With the rapid advancement in machine learning, SR techniques have been continuously evolving and successfully applied to various domains. 
For instance, they have been utilized for enhancing Computed Tomography (CT) images in medicine (e.g., \cite{CTSRCNN}), improving the resolution of electron microscopy images in material science (e.g., \cite{EMSR}), and addressing the challenges of enhancing precipitation data in the field of climatology, which is our focus. 
In recent studies, neural networks have been shown that they are able to skillfully generate satisfying climate downscaling results (e.g., \cite{DeepSD, FSRCNN-ESM, ResLap, climate_srgan01, climate_srgan02}). 
Like the training manner in the image field, in the above approaches, the input is the downsampled or degraded data from its corresponding ground truth. 
Here we use ``homogeneity'' to describe such training pairs for example in Fig. \ref{fig:homoheterdataset}a.
On the contrary, the term ``heterogeneity'' is used to describe the paired data from difference resources as shown in Fig.\ref{fig:homoheterdataset}b. 
In our study, we are more interested in applying super-resolution to heterogeneous data pairs and the data we use will be introduced in Sec. \ref{sec:dataset}. 
The heterogeneity problem in super-resolution is more realistic, as there is always a presence of bias between the simulation results and the observations. 
Therefore, it is crucial to account for and to correct the bias when performing super-resolution. 

Although an encoder-decoder network with residual learning is proposed to tackle this climate downscaling problem in a continental region (e.g. \cite{YNet}), a new architecture is required for areas with strong climatic forcings. 
Consequently, we are here to present a newly designed architecture for producing high-quality super-resolution results of precipitation data in such areas.
%% ------------------------------- Strong forcing areas ------------------------------------
Different from continental regions, island regions or countries often have more complex climate circumstances because they have stronger and more changeable forcings. 
Besides, these regions could have limited water resources because of insufficient areas for preserving water. 
Therefore, studying the long-term trends of precipitation in these regions is a challenging and active research topic in climatology (e.g., \cite{islandCyprus, islandIreland, islandJava}). 
For example, Taiwan is a subtropical island with only an area of 36,000 $km^2$, which is only 0.4\% of China or the United States. 
It has strong climatic forcings: the Siberian High and the Pacific Ocean ironical maritime air form the Stationary Front, southwesterly flows form convective rains in summer, and northeast monsoons bring orographic rains in winter. 
To more precisely capture the precipitation trend in Taiwan, a huge computational cost is required, which is not cost-efficient. 
Conventionally, climatologists in Taiwan simulate the long-term precipitation with low resolution and apply climate downscaling methods iteratively to obtain high-resolution results. 
However, it is still not likely to generate high-quality and realistic outputs from the low-resolution simulations due to unique geographic characteristics and uneven rainfall distribution in Taiwan.

%% ----------------------------------------- Our Model ----------------------------------------------
To address the challenges of climate downscaling and bias correction for the small-scale region, we proposed a deep neural network (DNN) consisting of three main parts: convolutional layers, attention blocks, one-step upscaling layer, and additional auxiliary data (topography data) concatenation shown in Fig.\ref{fig:ModelOverview}. 
Convolutional layers are responsible for bias correction and mapping the input low-resolution data to latent vectors. 
Attention blocks help to optimize the learning process (e.g., \cite{CBAM}). 
The one-step climate downscaling layers combine the learned feature maps and in our approach, we use a pixel shuffle (or pixel rearrangement) layer for image upscaling (e.g., \cite{ESPCN}). 
% If the climate downscaling factor is less than 8 times, our approach suggests a one-step climate downscaling scheme \cite{FSRCNN}. 
% Else, for the climate downscaling factor greater than 8 times, we suggest a cascading manner of staged magnification, which will provide better downscaling performance for a very-large scaling factor.
Moreover, \cite{topoinflu} have established that topography has a significant impact on precipitation. 
Therefore, our model takes the high-resolution topographical data as the auxiliary data to facilitate the high-quality climate downscaling results.

%%------------------------------------------ Contribution -------------------------------------------
Our main contributions are concluded below:
\begin{itemize}
    \item We propose a deep learning model for heterogeneous precipitation simulation data in climate downscaling problems with bias correction.

    \item The model is specially designed for an area with strong regional scale forcings and can receive a very small number of precipitation data points from the simulation and generates a corresponding high-resolution output. 

    \item  We conducted a comprehensive study and compared it with different types of climate downscaling approaches, including statistical methods and other machine/deep learning approaches to show that ours outperforms alternatives. 
\end{itemize}

% Homogeneity and heterogeneity of the dataset
\section{Related Works}
\label{sec:relatedwork}
% So far, we categorize downscaling methods into three classes: Statistical downscaling (SD), Dynamic downscaling, and machine learning-based climate downscaling methods. 
% Let us briefly introduce these downscaling methods and list some representative works below.
In this section, we categorize climate downscaling methods into three classes and discuss below: statistical, dynamic, and machine learning based climate downscaling methods. 
Also, we briefly introduce the super-resolution models in the image field. 

%%%%%%%%%%%%%%%%%%%%%%%%%%%%%%%%%%%%%%%%%%%%%%%%%%%%%%%%%%%%%%%%%%%%%%%%%%%%%%%%
    \subsection{Statistical Climate Downscaling}

    \cite{stat01} conducted statistical climate downscaling (SD), relying on statistical correlations between General Circulation Models (GCMs) and regional historical observational data.
    It assumes that the relationship is constant and defines predictand as a function of the predictor. 
    This type of approach can be divided into two steps: empirically establish the statistical relationship between the simulation results and local variables (e.g., \cite{stat02, stat03}), and then apply this relationship to the simulation results to map to local-scale climate variables, including precipitation, air pressure, or temperature.

    There are many ways to perform SD tasks.
    For example, regression models such as the generalized linear model (e.g., \cite{GLM}) have been widely used to fit probability distributions (e.g., \cite{glm01, glm02, glm03}). 
    The non-homogeneous Hidden Markov Model (NHMM), introduced by \cite{NHHM}, tries to relate rainy occurrences (e.g., \cite{nhhm01, nhhm02, nhhm03, nhhm04}).
    Two SD methods, Quantile Mapping (QM) and Bias Corrected Spatial Disaggregation (BCSD) researched by \cite{BCSD}, are widely used in coupling GCMs and historic observations. 
    QM tries to derive corrected precipitation values by matching the cumulative distribution functions (CDFs) of the biased one and non-biased one (e.g., \cite{QM}). 
    BCSD further optimizes QM by adding a Spatial Disaggregation step, where the corrected values are interpolated based on the average of the corresponding days, rather than themselves.
    However, SD methods can only get the spatially finer data. 
    Since the original data are GCM outputs, they do not consider the detailed influences of the landform, soil, or vegetation. 
    In other words, regional climate mechanisms are omitted during the process. 
    Therefore, as the scaling factor increases, the biases become larger and cause spatial shifts on the grid points.

    % --------------------------------------- RCMs ---------------------------------------
    \subsection{Dynamic Climate Downscaling}
    Dynamic climate downscaling, or the Regional Climate Models (RCMs), are similar to GCMs. 
    Both are formulated on physical principles, but RCMs focus on a smaller, local scale climate simulations rather than the global ones. 
    ``Dynamic'' represents the interactions between physic and chemical variables of RCMs. 
    Due to such environmental interactions, the main downside of this approach is the huge computational costs. 
    To generate regional predictions, scientists will feed RCMs with GCMs simulation results as the boundary conditions. 
    Significantly, different boundary condition settings generate distinguishing climate patterns, so do the area size and resolution, and each setting gives limited climate patterns in local. 
    Moreover, with limited resources, there is a trade-off between spatial or temporal resolution. 

    Although RCMs have powerful comprehension upon local features, there are still systematic errors, leading to large uncertainties (e.g., \cite{RCMerr}). 
    To correct such errors, some ensemble methods are proposed for improving RCMs, but they require collaboration of institutions due to the intensive computation and intergration (e.g., \cite{rcm01, rcm02, rcm03}).

%%%%%%%%%%%%%%%%%%%%%%%%%%%%%%%%%%%%%%%%%%%%%%%%%%%%%%%%%%%%%%%%%%%%%%%%%%%%%%%%
    \subsection{Single-Image Super-Resolution with CNNs}
    Single-image super-resolution (SISR) is a classic application in the image processing field by \cite{SISRreview}. 
    The popularity of deep learning has skyrocketed in recent decades, leading to the successive proposal of various architectures for SISR. 
    Among these models, convolutional neural networks (CNNs) are the most popular one.

    % --------------------------------------2014 SRCNN ---------------------------------------
    \cite{SRCNN} proposed the very first end-to-end CNN-based SISR model, known as Super Resolution Convolutional Neural Network (SRCNN). 
    Subsequently, \cite{FSRCNN} also proposed an improved version of SRCNN, named Fast Super-Resolution Convolutional Neural Networks (FSRCNN), which can perform SISR without the bicubic upscaling process in SRCNN. 
    % Successively, various SR models are proposed, using convolutional layers as the spine of the model and adding other techniques to achieve higher performance and better output quality.
    % ----------------------------------2016 ESPCN, FSRCNN -----------------------------------
    \cite{ESPCN} used the pixel rearrangement (pixel shuffle) method as the image upscaling layer to build Efficient Sub-pixel Convolutional Neural Networks (ESPCN). 
    It uses a one-step image upsampling layer which reduces the number of model parameters while maintaining reliable performance. 
    % -----------------------------------2016 VDSR, DRCN ------------------------------------
    % Kim et al. proposed Very Deep Super Resolution networks (VDSR)\cite{VDSR} which features the residual learning (skip connections) \cite{ResNet} and gradient clipping for mitigating the gradient vanishing problem \cite{GradientClipping}. 
    % And it also extends to be a multi-scale-factor model. 
    \cite{VDSR} proposed Very Deep Super Resolution networks (VDSR) that features residual learning (skip connections) (e.g., \cite{ResNet}). 
    Another work called Deeply-Recursive Convolutional Network (DRCN) introduced by \cite{DRCN}, is the first to apply the Recursive Neural Network (RNN) architecture to SR problems.
    % Specially, DRCN combines intermediate results from different recursion blocks to generate a better final prediction. 
    % And we also implement similar behavior in our model which will be elaborated in \cref{sec:method}. 
    % -------------------------------------- 2016 REDNet ------------------------------------
    \cite{REDNet} proposed Residual Encoder-Decoder Network (REDNet) which contains an architecture of convolutional (encoders) and deconvolutional (decoders) layers with symmetrically multiple skip connections, which is the first approach in image restoration by using these two layers. 
    % Convolutional layers are used for feature extraction and encode the main component of images, while deconvolutional layers decode the image abstraction and are used for image recovery. 

    % -------------------------------------- 2017 LapSRN ------------------------------------
    In addition, \cite{LapSRN} applied a cascade of convolutional layers for feature extraction and deconvolutional layers for image upscaling. 
    It features in progressive reconstruction of SR predictions, using Laplacian pyramids. 
    % Their model, named LapSRN, offers much flexibility and applicable to resource-aware adaptability. 
    % -------------------------------------- 2017 SRGAN ------------------------------------
    \cite{SRGAN} set a new state-of-the-art on public benchmark datasets by introducing the Generative Adversarial Network (GAN) architecture for SISR. 
    It borrows the VGG layers from \cite{vgg} to replace commonly used loss function such as MSE, and defines a so-called VGG loss which is like a perceptual similarity function between feature representations and generated images. 
    % Many of the above architectures are extended to climate downscaling applications. 

%%%%%%%%%%%%%%%%%%%%%%%%%%%%%%%%%%%%%%%%%%%%%%%%%%%%%%%%%%%%%%%%%%%%%%%%%%%%%%%%%%%%%%%%%%%%%%%%%% 
    \subsection{Deep Learning Based Climate Downscaling}
    % In recent years, more and more climate models adopt super-resolution techniques to downscale satellite or remote sensing data in different levels for various purposes. 
    % Most of them are inspired from those in SISR field. 
    % The main difference from image and climate data is the theoretically unlimited minimum and maximum values of the channels, or parameter values. 
    % Fortunately, with some modifications and decorations in normalization and modules, it once again shows the capabilities of deep learning. 
    In recent years, more and more climate models have adopted super-resolution techniques in SISR field to climate datasets for further analysis. 
    The main difference from image and climate data is the theoretically unlimited minimum and maximum values of the channels, or parameter values. 
    Fortunately, with some modifications, it once again shows the capabilities of deep learning.

    % ------------------------------------- 2017 DeepSD ------------------------------------------
    % DeepSD \cite{DeepSD} is one of the earliest deep learning model trying to downscale the climate data. 
    % It is modified from SRCNN, but in a stacked manner. 
    % For each downscale factor, the model blocks are trained independently with its associated training data pairs. 
    % Different scales of elevation are added as additional channel to the precipitation inputs. 
    % For instance, breakdown a scale factor of 8 into three $\times$2 blocks. 
    % It also takes the elevation as input, concatenated with the low-resolution pieces. 
    % DeepSD has set a baseline of downscaling method in application of machine learning in climatology. 
    % It is still applied in recent years for easy training and fast outputs \cite{deepsdindia}. 
    % Sharma et al. \cite{ResDeepD} also add skip connections to SRCNN blocks in a residual learning manner. 
    DeepSD, proposed by \cite{DeepSD} is one of the very first deep learning models that tries to perform the climate downscaling of precipitation data. 
    It is modified from SRCNN, but in a stacked manner. 
    For example, divide a scale factor of 8 into three $\times$2 blocks. 
    It also takes the topography data as an input channel, taking the geographical influences into account. 
    Some variants of DeepSD are proposed for easier er training and faster outputs (e.g., \cite{deepsdindia, ResDeepD}).
    % ------------------------------------ 2022 FSRCNNESM ------------------------------------------
    Beyond precipitation data, \cite{FSRCNN-ESM} tried training a modified FSRCNN model, named FSRCNN-ESM, which takes earth system model simulations as input. 
    They added additional convolutional layers after the deconvolution step in the FSRCNN to improve the climate downscaling performance. 
    % The inputs include five variables: surface temperature , shortwave heat flux, long-wave heat flux, precipitation convective rate, and the large scale precipitation rate. 
    % Interestingly, they also took elevation data as second input channel, but found no such improvement in their case. 
    % Yet, the same concern is that the image upscaling step in SRCNN-based model is performed at the very beginning of the layers. 
    % As the upscale factor gets larger, it gives rise to the longer computational time and re-training. 

    % ------------------------------------ 2019 ResLap ------------------------------------------
    \cite{ResLap} introduced a residual dense block to LapSRN network.
    One can collect different scales of high-resolution results at corresponding level. 
    They also conducted a detailed study of checkerboard artifacts elimination in parameter studies of deconvolutional layers. 

    % ------------------------------------ 2020, 2022 SRGAN -------------------------------------
    Similar to the progress in SISR field, one of the latest climate downscaling methods is SRGAN-based model which turns out to outperform other types of models like DeepSD, Augmented Convolutional Long Short Term Memory (ConvLSTM), LapSRN, and U-Net (e.g., \cite{climate_srgan01, climate_srgan02}). 
    
    However, all models above are trained with homogeneous pairs, that is, degraded one as input, and identical one as ground truth. 
    As mentioned in Sec. \ref{sec:intro}, we use the term, homogeneity, as that there are a few differences between the input (low resolution) and output (high resolution), meaning that we can still tell the patterns or structures by traditional interpolations like bilinear or bicubic. 
    On the contrary, we are trying to construct a model to capture the relationships between the reanalysis precipitation data and the corresponding observation data which are obviously not from the same data source, saying a heterogeneous pair. 
    % -------------------------------------- 2020 YNet ------------------------------------------
    Addressing the heterogeneous dataset, YNet, as developed by \cite{YNet}, used GCMs simulations and reanalysis data for the training. 
    % Inspired by REDNet, YNet also forms in symmetrical convolution-deconvolution layer pairs, or the encoder-decoder pairs.
    % To avoid chess-board, it adds a convolutional layer after each deconvolutional one. 
    % It shows better performance than DeepSD in metrics of MSE, bias, and Pearson correlation coefficient.
    Yet, YNet tackled this climate downscaling problem in a continental region, which has a weak climatic forcings. 
    Therefore, we have to consider a new architecture to cover both heterogeneity and strong climatic forcings which will be elaborated in Sec. \ref{sec:method}. 

\section{Methods}
\label{sec:method}
The main purpose of this work is to get a high-resolution result \textbf{Y}, given a low-resolution input \textbf{X}, which can be simplified as: $\mathbf{Y} = Model(\mathbf{X};\Theta)$, where $\Theta$ is the hyperparameters of the model. 
Our model comprises a series of convolutional layers with residual attention blocks as our model spine, skip connections of feature maps at different levels, and a one-step image upscaling layer, as shown in Fig. \ref{fig:ModelOverview}. 
These parts are introduced in the following subsections.

%%% KC %%%
% The main purpose of this work is to train a deep learning model to predict high-resolution precipitation data \textbf{Y} from a given low-resolution input \textbf{X}. This process can be represented as $\mathbf{Y} = Model(\mathbf{X};\Theta)$, where $\Theta$ is the hyperparameters of the model. 
% Our model comprises a series of convolutional layers with RABs as the model spine, forward connections of attention maps at different levels and one connection of traditional interpolation, and a one-step image upscaling layer, as shown in \cref{fig:ModelOverview}. 
% These parts are introduced in the following subsections.

% ======================================== Sequential CNN ==================================================
    \subsection{Cascading Convolutional Bias Correction}
    We utilize a cascade of convolutional layers to fulfill the tasks of bias correction and feature extractions as shown in Fig. \ref{fig:ModelOverview}a. 
    Given a low-resolution input \textbf{X} $\in\mathbb{R}^{H\times W\times C}$, a convolutional layer outputs a feature map \textbf{F} $\in\mathbb{R}^{H\times W\times f}$, where \textit{H}, \textit{W}, and \textit{C} are height, width, and the number of input channels and $f$ is the number of filters, respectively. 
    It can be formulated as Eq.(\ref{eq:convd}): 

    \begin{equation}
        \label{eq:convd}
        \begin{split}
            \mathbf{F}_0 & = act(W \ast \mathbf{X} + bias), \; and \\
            \mathbf{F}_i & = act(W_i \ast \mathbf{F}_{i-1} + bias_i) \qquad \forall \;i \in \mathbb{N}^{+}
        \end{split}
    \end{equation}

    for subsequent layers, where $\mathbf{F}_{i}$ is the intermediate feature map of $i^{th}$ layer, $(\ast)$ denotes the convolution operation, and $act(\cdot)$ as the activation function. 
    We can see there is a bias term in the operation, exactly what we expect to be analogous to statistical climate downscaling methods (QM), where the CDF and its inverse are replaced by a nonlinear mapping, as shown in Eq.(\ref{eq:convd_analogous_to_qm}). 

    %%%KC%%%
    % One of the main tasks of this work is to correct the bias in the input data. The concept has been illustrated in Fig. \cref{1b}. 
    % We design a cascade of convolutional layers to fulfill this requirement and also to perform feature extractions \cref{fig:ModelOverview} (a). 
    % Given a low-resolution input \textbf{X} $\in\mathbb{R}^{H\times W\times C}$, a convolution layer outputs an feature map \textbf{F} $\in\mathbb{R}^{H\times W\times f}$, where \textit{C} is the number of input channels and $f$ is the number of filters. 
    % It can be formulated as \cref{eq:convd}: 
    % \begin{equation}
    %     \label{eq:convd}
    %     \begin{split}
    %         \mathbf{F}_0 & = act(W \ast \mathbf{X} + bias), \; and \\
    %         \mathbf{F}_i & = act(W_i \ast \mathbf{F}_{i-1} + bias_i) \qquad \forall \;i \in \mathbb{N}^{+}
    %     \end{split}
    % \end{equation}
    % for subsequent layers, where $\mathbf{F}_{i}$ is the intermediate feature map of the $i^{th}$ layer, $\ast$ denotes the convolution operation and $act(\cdot)$ as the activation function. 
    % In \cref{eq:convd}, there is a bias term in the operation, that is exactly what we expect to be analogous to \cref{eq:QM}, where the CDF and its inverse are replaced by a nonlinear mapping, as shown in \cref{eq:convd_analogous_to_qm}. 

    \begin{equation}
        \label{eq:convd_analogous_to_qm}
        x_{i,j}^{cor} = W_{n} \ast (\mathbf{F}_{n-1}(...(\mathbf{F}_{0}(x_{i,j}^{bias})...))) + bias_n
    \end{equation}
% ======================================== Skip Connections ==================================================
    \subsection{Skip Connections}
    To avoid gradient diminishing for training a DNN, skip connections have been widely used for both residual learning and decreasing training parameters. 
    
\begin{figure}[tb]
    \centering
    \includegraphics[width=0.4\columnwidth]{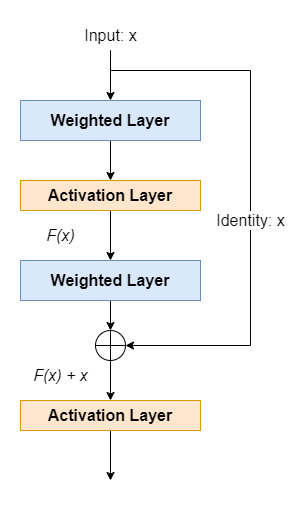}
    \caption{A simple building block with a skip connections. 
    The skip connection (or ``shortcut connection'') is to feed forward an identity $x$ to the output of a series of layers. 
    It is commonly used in a deep neural network for training optimization.}
    \label{fig:skipconnect}
\end{figure}

\begin{figure*}[tb]
    \centering
    \includegraphics[width=\textwidth]{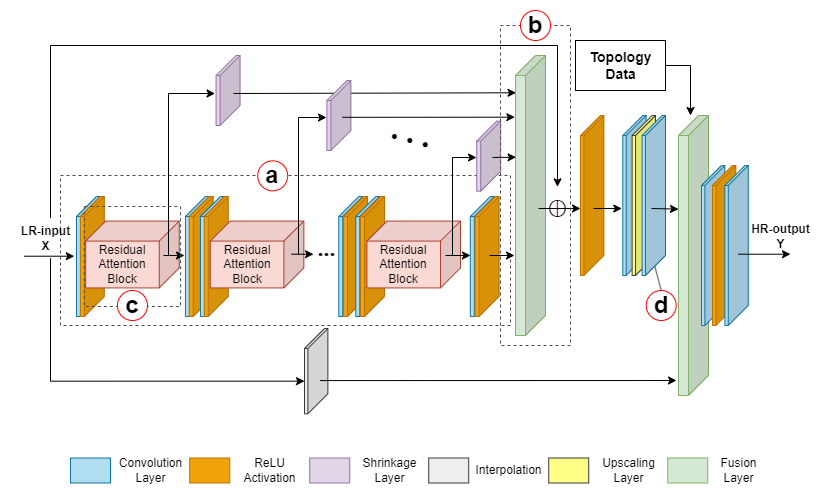}
    \caption{Overview of proposed model architecture. 
    Main stream of the model is a series of residual attention block, attached with a convolutional layer and a ReLU activation. 
    Refined intermediate feature maps are fed into a convolutional layer with 1$\times$1 kernel size to limit the number of parameters and pick up the most significant one at different levels for forwarding. 
    The first fusion layer combines all refined feature maps, while the second one additionally concatenate elevation data and interpolation of original input. 
    The image upscaling layer is adopting the pixel shuffle technique from ESPCN, after a convolutional layer with the number of filters equal to square of scaling factor.}
    \label{fig:ModelOverview}
\end{figure*}

    As shown in Fig. \ref{fig:ModelOverview}b, let $\mathbf{F}(x)$ be the desired mapping of input $x$, and the residual learning is reforming the nonlinear mapped into $\mathbf{F}(x)-x$ (Fig. \ref{fig:skipconnect}). 
    It is said to be easier in optimizing the residuals than original mapping. 
    To a extreme that the mapping is identical, to learn a zero residual is easier than to fit the original one by stacked non-linear layers (e.g., \cite{ResNet}). 
    Inspired from \cite{RDB}, to make the most use of the feature maps extracted from previous layers, and to avoid gradient diminishing, we also apply several skip connections in our model in Fig. \ref{fig:ModelOverview}b. 
    We extract the feature maps from convolutional layers and apply a shrinkage layer for lessening the number of parameters.
    These extracted maps are then connected to a fusion layer before climate downscaling.

    %%%KC %%%
    % [~~We need a sentence at the beginning to motivate why we need the skip connection~~]
    % To deal with this potential problem, we add the skip connection scheme to our deep learning architecture. \cref{fig:skipconnect} is the illustration of skip connection. Assume that $\mathbf{F}(x)$ is the desired mapping of the input $x$, then residual learning is the reformulation of the nonlinear mapping into $\mathbf{F}(x)-x$. With the skip connection, it is easier to optimize the residuals than with the original mapping. Even if the mapping is identical, learning a zero residual is easier than learning fitting the original one by stacked non-linear layers \cite{ResNet}. Also, inspired by Zhang et al. \cite{RDB}, to make use of the feature maps extracted from previous layers, and to avoid gradient diminishing, several connections are applied in our model in \cref{fig:ModelOverview}(b).  Our model extracts feature maps from convolutional layers and applies a shrinkage layer to reduce the number of parameters.
    % These extracted maps are then connected to a fusion layer before climate downscaling.

% ======================================== Residual Attention Block ==================================================
    \subsection{Residual Attention Block}
    We adopt the Channel Block Attention Module (CBAM) proposed by \cite{CBAM} in Fig. \ref{fig:ModelOverview}c. 
    CBAM has two sub-modules: Channel Attention Block (CAB) and Spatial Attention Block (SAB). 
    These two blocks help our model to learn ``what to emphasize'' and ``where to focus'' respectively, and emphasize the extracted features by multiplying to input feature map. 
    % More specifically, they are meant to emphasize meaningful patterns and to suppress unnecessary ones. 
    
\begin{figure}[tb]
    \centering
        % RAB
        \begin{subfigure}[t]{0.9\columnwidth}
            \includegraphics[width=\columnwidth]{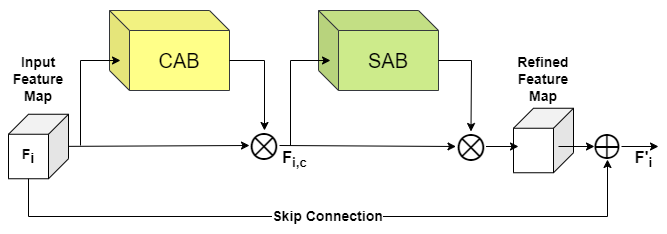}
            \caption{Residual attention block}
        \end{subfigure}
        \hfill
        % CAB
        \begin{subfigure}[t]{0.45\columnwidth}
            \includegraphics[width=\columnwidth]{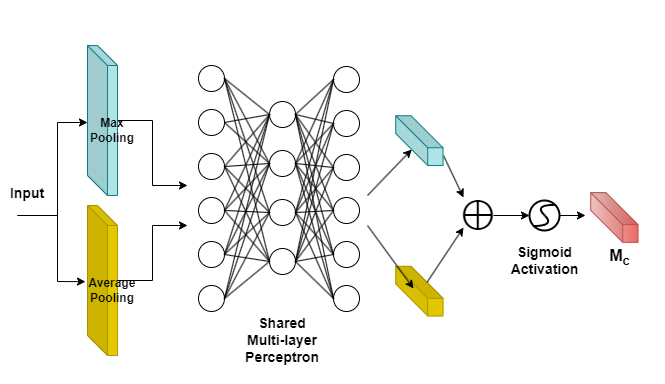}
            \caption{Channel attention block}
        \end{subfigure}
        % SAB
        \begin{subfigure}[t]{0.45\columnwidth}
            \includegraphics[width=\columnwidth]{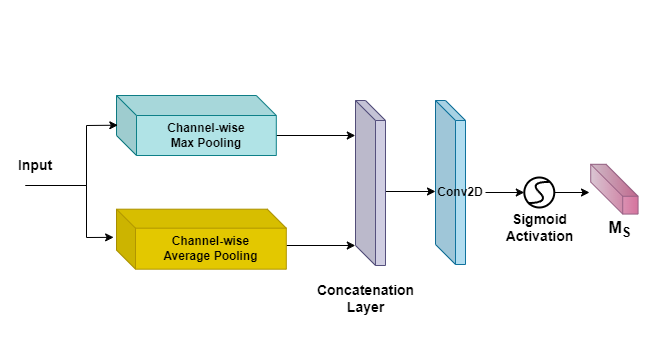}
            \caption{Spatial attention block}
        \end{subfigure}
    \caption{Diagram of attention blocks. 
            As illustrated in (a), a residual attention block (RAB) consists of a channel attention block (CAB) and a spatial attention block (SAB), and a skip connection from input feature map forwarding to refined feature map. 
            (b) shows the structure of CAB, including a global max pooling, a global average pooling, and a shared multi-layer perceptron unit. 
            (c) illustrates the structure of SAB, including a pair of channel-wise max and average pooling, a concatenation layer, and a convolutional layer. Both activation functions in (b) and (c) are using sigmoid activation.}
    \label{fig:CBAM}
\end{figure}

    CAB consists of a pair of max-pooling and average pooling, and a shared multi-layer perceptron (MLP) unit, as shown in Fig. \ref{fig:CBAM}b. 
    SAB consists of a pair of channel-wise max-pooling and average pooling and a convolutional layer, as shown in Fig. \ref{fig:CBAM}c. 
    Given an intermediate feature map of the i-th layer $\mathbf{F}_i \in\mathbb{R}^{H\times W\times f}$, CAB gives a 1D attention map $M_{c} \in\mathbb{R}^{1\times 1\times f}$, while a 2D one $M_{s} \in\mathbb{R}^{H\times W\times 1}$ is generated by SAB. 
    The refined feature maps $\mathbf{F}_{i,c}$ and $\mathbf{F}_{i,s}$, are summarized as Eq.(\ref{eq:CBAM}): 

    \begin{equation}
        \label{eq:CBAM}
        \begin{split}
            F_{i,c} & = M_{c} \otimes \mathbf{F}_i \\
            F_{i,s} & = M_{s} \otimes \mathbf{F}_i,    
        \end{split}
    \end{equation}
    
    where $\otimes$ denotes the element-wise multiplication. 
    For $M_{c}$, it is broadcasted along spatial dimension.
    We found that it has better performance when arranging in a sequential manner, and we added a connection from $\mathbf{F}_i$ to construct the so-called Residual Attention Block (RAB) shown in Fig. \ref{fig:CBAM}a. 
    The final refined feature map $\mathbf{F}_{i}^{\prime}$ is shown in Eq.(\ref{eq:CBAM_seq}):
    \begin{equation}
        \label{eq:CBAM_seq}
        \mathbf{F}_i^{\prime} = \mathbf{F}_{i} \oplus (M_{s} \otimes F_{i,c}),
    \end{equation}
    where $\oplus$ denotes the element-wise addition.
    \subsection{One-step Image Upscaling Layer}
\begin{figure}[tb]
    \centering
    \includegraphics [width=0.8\columnwidth]{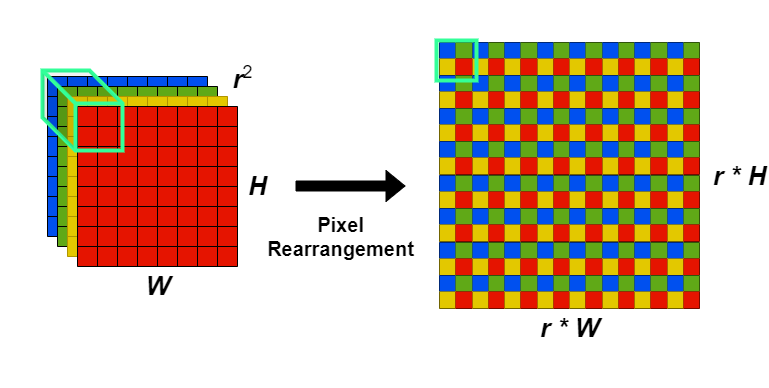}
    \caption{Pixel shuffle from ESPCN. 
            This one-step upscsaling layer rearranges the pixels along channel axis into a larger image. 
            The number of channels needs to be adjusted to the square of the scaling factor. 
            One can change the number of filters of the convolutional layer right before pixel shuffle layer for preferred scaling factor.
            }
    \label{fig:pixelre}
\end{figure}

    We use the pixel shuffle method from ESPCN (Fig. \ref{fig:pixelre}) and attach another convolutional layer to mitigate the blocky results as shown in Fig. \ref{fig:ModelOverview}d. 
    Pixel shuffle, or pixel rearrangement, is to push the subpixels along the channel axis to the spatial field. 
    For example, an intermediate output with $r^2$ number of filters is fed into pixel shuffle layer which gives a $rH \times rW$ upscaled results. 
    It has more modeling capabilies than other image upscaling methods. 
    Others such as bilinear, bicubic and deconvolutional layer may generate blurred images or severe checkerboard artifacts. 
    Bilinear and bicubic interpolation produce blurry results and do not consider spatial relationship between grid points, which may result in distorted outputs. 
    While in deconvolutional layer, the uneven kernel overlapping issue occurs, resulting in checkerboard artifacts especially when its kernel size is not divisible by the stride (e.g., \cite{deker}). 
    We will also compare the performance of different image upscaling layer in Sec. \ref{sec:exp}.

    %%% KC %%%
    % To downscale and produce the final high-resolution result, we use the ESPCN pixel shuffle method \cref{fig:pixelre} and attach another convolutional layer to mitigate blocky results. These components are shown in \cref{fig:ModelOverview} (d). 
    % Pixel shuffle, or pixel rearrangement, is used to reorganize the subpixels along the channel axis to the spatial field. 
    % An intermediate output with $r^2$ filters is fed into the pixel shuffle layer, which can produce a $rH \times rW$ upscaled result. 
    % This module has more modeling power than other upscaling methods, such as deconvolution upscaling and traditional interpolation. 
    % Other upscaling layers such as bilinear, bicubic, and deconvolutional layers may generate blurred images or result in severe checkerboard artifacts. 
    % Bilinear and bicubic interpolation produce blurry results because they do not consider spatial relationships between grid points, resulting in distorted outputs. 
    % And the uneven kernel overlapping issue occurs in the deconvolutional layers approach, resulting in checkerboard artifacts, especially when its kernel size is not divisible by the stride \cite{deker}. 
    % We will also compare the performance of different upscaling layer in \cref{sec:exp}.

% ======================================== Implementation ==================================================
    \subsection{Implementation Details}
    % ------------------------ log1p normalization ---------------------------
    We normalized the training data and elevation by log1p transformation: $x' = ln(x+1)$ to handle the long-tail distribution property of precipitation, and no batch-normalization during the training process. 
    % ------------------------ conv2d parameters ---------------------------
    In our model, all convolutional layers are set with 3${\times}$3 kernel size throughout our model backbone, along with 64 filters, strides by 1 and valid paddings (zero paddings), except a 5$\times$5 kernel size in SAB, and a 1${\times}$1 kernel size in shrinkage layers. 
    % ------------------------ deconv2d parameters ---------------------------
    % A kernel size of 5${\times}$5, strides by 5, is set for deconvolutional layer as image upscaling layer \cite{deker}.
    % ------------------------ activation function ---------------------------
    A rectified linear unit (ReLU), proposed by \cite{relu}, is used as an activation function after each convolutional layer except the last one and the one in the attention blocks, where a sigmoid activation is applied.
    
    % ------------------------ skip connections ---------------------------
    We extract intermediate feature maps from every two convolutional layers into the attention block, which consists of a pair of sequentially arranged CAB and SAB.
    % ------------------------ CBAM paramters ---------------------------
    The shared MLP of CAB is set to be 256 nodes, a reduction factor of 0.5, and the output nodes are equal to the number of channels, in a total of three dense layers. 
    % The CAB and SAB are arranged in a sequential manner. 
    The RAB outputs are then fed into shrinkage layers. 
    Subsequently, we concatenate them and feed them into a fusion layer, obtaining an output channel of 1 for the element-wise addition with skip connections from the local input, namely the residual learning mechanism. 
    
    % ------------------------ downscaling layer -----------------------------
    Next, we feed into a pixel shuffle layer for climate downscaling. 
    Notice that we add a convolutional layer immediately after it to alleviate checkerboard artifacts. 
    % ------------------------ elevation fusion -----------------------------
    Eventually, we append several convolutional layers again for the elevation data fusion to generate the final result of climate-downscaled precipitation data.

% ---------------------------------------- Figures ----------------------------------------------
% skip connection

% Pixel shuffle

% Residual Attention Block

% Main stream of model
\section{Dataset}
\label{sec:dataset}
Continuing with the heterogeneity and strong climatic forcings, we found Taiwan is a suitable study area to showcase our model capabilities. 
Taiwan is an island in subtropics, extending about 400 km from north to south and 150 km from east to west with only an area of 36,000 km$^2$, about 4‰ large of China. 
Mountains run along north-south manner and most of rivers flow east to west. 
The monsoon season starts from April to October, mainly contributed by plum rains in May to June and typhoon rains in July to September. 
While in northern cities, northeast monsoon also causes orographic rain in winter.
Despite the average annual rainfall being up to 2,500 mm, which is 2.5 times than the global average, the water resource is insufficient, even lower than the global average. 
Such insufficiency is attributed to geographical characteristics and uneven rainfall distribution throughout the year.

%%% KC %%%
% We will use Taiwan precipitation datasets to carry out our experiments, so we will introduce the datasets in detail in this section.
% Taiwan is an island in the subtropics, extending approximately 400 km from north to south and 150 km from east to west with only an area of 36,000 $km^2$, about 4‰ large of China or the United States. 
% The mountains run north to south, and most of the rivers flow east to west. 
% The monsoon season starts from April to October, mainly contributed by plum rains from May to June and typhoon rains from July to September. 
% In northern cities, the northeast monsoon also causes orographic rain in winter.
% Although the average annual rainfall is up to 2,500 mm, which is 2.5 times the global average, the water resource is insufficient, even less than the global average. 
% This insufficiency is attributed to geographical characteristics and uneven rainfall distribution throughout the year. 

% This essentially raises the storage and preservation of water resources. 
% Aside from water saving policies, governments are cooperating with climatologists in hydrological plannings. 

We are taking ERA5 reanalysis data as our training input, pairing with TCCIP observations as ground truth, and train our model in a supervising manner. 
The reason is that since the upscaling factor is minimal for simulations to downscale into observations in Taiwan area, we are not using GCMs or RCMs as our model input, or the upscaling factor would be up to nearly a hundred. 

%%%KC %%%
% We are taking ERA5 reanalysis data as the low-resolution input and TCCIP observations as the high-resolutin datasets.
% [~~~I think we can add the training dataset and testing dataset description here ~~~]
% ERA5 is the reanalysis results of the GCM simulation and the historical data set, and Taiwan's climatologists often use it to study long-term climate change in Taiwan. 

The following subsections introduce the attributes and dimensions of the data:
    \subsection{ERA5 Reanalysis Data}
    ERA5 dataset is produced using Four-Dimensional Variational Data Assimilation (4D-Var) and model forecasts of the European Center for Medium-Range Weather Forecasts (ECMWF) Integrated Forecast System (IFS). 
    It is a kind of reanalysis data, which combines a large number historical observations into global simulation estimates. 
    \defcitealias{ERA5}{ERA5}
    Climatologists often adopt ERA5 dataset for research to understand climate change and current weather extremes (website: \citetalias{ERA5}). 
    The spatial unit of ERA5 dataset is represented in meters, and the time span is from the 1950 to 2020, hourly. 
    For each daily piece, the global precipitation values are stored in a matrix with the size of 1440$\times$720 along longitude and latitude.

    \subsection{TCCIP Precipitation Data}
    We take gridded precipitation data from \cite{TCCIPgrid} as our high-resolution data, which is from the Taiwan Climate Change Projection Information and Adaptation Knowledge Platform (TCCIP). 
    The dataset is rasterized based on the observatory data, collected from Taiwan Central Weather Bureau, Water Resources Agency, and Civil Aeronautics Administration. 
    There are two different spatial resolutions available: 0.05$^{\circ}$ and 0.01$^{\circ}$. 
    The 0.05$^{\circ}$ one means that the distance between neighboring data points is around 5 km, covering from latitude 21.9$^{\circ}$N to 25.2$^{\circ}$N and longitude 120.0$^{\circ}$E to 122.0$^{\circ}$E, with the size of 69$\times$41. 
    While 0.01$^{\circ}$ one means that the distance between neighboring data points is around 5 km, covering from latitude 21.9$^{\circ}$N to 25.2$^{\circ}$N and longitude 120.0$^{\circ}$E to 122.0$^{\circ}$E, with the size of 69$\times$41. 
    Data are recorded daily, from 1960 to 2020, 22,281 days in total, and the precipitation unit is in millimeters. 
    More information is available at the \cite{TCCIP} website. 
    % The detailed raterization steps is performed as the following steps: 
    % \begin{enumerate}[1)]
    %     \item Apply Gaussian Process Latent Variable Model (GPLVM) \cite{GPLVM} method onto observatory data for normalization and parametrization by month and convert into Latent Gaussian Variable, which is normally distributed \cite{GPLVM2, GPLVM3}.
    %     \item Find optimized statistical model and parameters for each month and impute missing values \cite{Impute}.
    %     \item Decompose data into temporal and spatial parts by Empirical Orthogonal Function (EOF) and after keeping 99.5\% variation, we rasterize the eigenvectors and the averages of excluded EOF values by Natural Neighbor Interpolation method \cite{NNI}.
    %     \item Multiply the principle components obtained from EOF analysis with the rasterized eigenvectors and add to each cell average value to get normalized gridded precipitation data.
    %     \item Substitute the values in step.4 into the model in step.2 to denormalize precipitation data, the gridded observation data with 1km resolution. As for the 5km resolution one, it is obtained from 1km one by Average Nearest Neighbor method.
    % \end{enumerate}

    \subsection{Topographical Data}
\begin{figure}[tb]
    \centering
        % Topology
        \begin{subfigure}[t]{0.45\columnwidth}
            \includegraphics[width=\columnwidth]{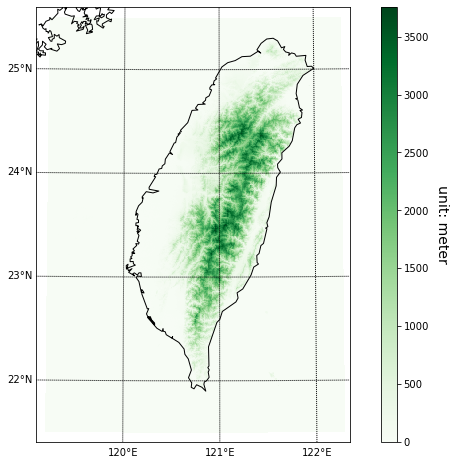}
            \caption{Topology of Taiwan island}
        \end{subfigure}
        \hfill
        % Divergence
        \begin{subfigure}[t]{0.45\columnwidth}
            \includegraphics[width=\columnwidth]{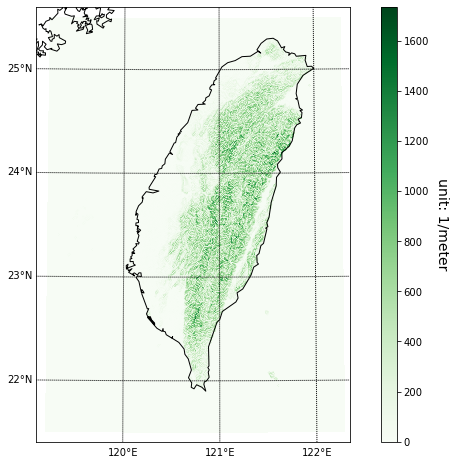}
            \caption{Divergence of elevation}
        \end{subfigure}
        \hfill
    \caption{Topology of Taiwan island. 
            (a) Study area in Taiwan. 
            Taiwan is an island in subtropics. 
            It extends about 400 km from north to south and 150 km from east to west. 
            Altitudes are shown in meters. 
            It shows the mountains run along north-south manner and therefore most of rivers flow from east to west. 
            (b) Drastic elevation changes in Taiwan. 
            The figure shows the divergence value of each data point, $\nabla_{x,y}$. By considering the neighboring altitudes along longitude and latitude, divergence is calculated as the square root of slopes in units of meters$^{-1}$.}
    \label{fig:topo}
\end{figure}

    Terrains are one of the most important factors in climatology, which affects air convection, lift, and flow. 
    For example, orographic precipitation is formed when moist air is forced to rise along the windward side of mountains and then becomes cooled, condensed, and eventually forms clouds. 
    On the contrary, the leeward side would encounter dry winds, even Foehn winds. 
    Therefore, our model also takes elevation information from the terrain to predict the high-resolution data. We use the terrain data from the Center for Geographic Information System, the Research Center for Humanities and Social Sciences, Academia Sinica as shown in Fig. \ref{fig:topo}. 
    The resolution of the data is 0.01$^{\circ}$ (1 km), with the resolution of 480$\times$371, latitude from 21.5042$^{\circ}$N to 25.4958$^{\circ}$N, and longitude from 119.2042$^{\circ}$E to 122.2875$^{\circ}$E. 
    The altitude range is from -36.930 to 3706.753 meters. 
    
    % The altitude characteristics are shown in \cref{fig:topodiv} by divergence $\nabla_{i,j}$ in \cref{eq:topodiv} where the boundary points are not calculated.
    
    \subsection{Data Preprocessing}
    To align with our TCCIP precipitation data (high-resolution data), ERA5 reanalysis data (low-resolution data) is converted to daily average and multiply with $10^3$ to get daily average precipitation data (unit: mm) with spatial resolution of 0.25$^{\circ}$ (around 25 km), and choose from the year 1960 to 2020, in a total of 22,281 time steps. 
    Furthermore, since we are only interested in the data points in Taiwan, we crop the data covering Taiwan island only, with a size of 14${\times}$9 per day, as low-resolution inputs.
    
    % The TCCIP gridded data is stored in one dimension as \textsl{csv} files, recording only on-land value. 
    For TCCIP data, we zero out the on-sea values and align the data points with the cropped ERA5 dataset near Taiwan island. 
    The TCCIP data is cropped from latitude 22.0$^{\circ}$N to 25.25$^{\circ}$N and longitude 120.0$^{\circ}$E to 122.0$^{\circ}$E by nearest neighboring interpolation. 
    Filtered data ends up with a resolution of 66${\times}$41 and each data point represents the precipitation data in a 5${\times}$5 km$^2$ region.
    % Precipitation values of 5km gridded dataset range from 0 to the maximums of 1,402 mm. 
    % We also did a little interpolations (bicubic, in this case) upon the 5km one to adjust the size of 70${\times}$45, which is exactly five times of resolution of ERA5 data.

    Regarding the terrain data, we also take the latitude and longitude with 22.0042$^{\circ}$N to 25.2542$^{\circ}$N, 120.0042$^{\circ}$E to 122.0042$^{\circ}$E, with the size of 391$\times$241, and mask out the negative values elevation data points (lower than sea level). 

% Taiwan Topology Divergence Equation
% \begin{equation}
%     \label{eq:topodiv}
%         \nabla_{i,j} = \sqrt{(x_{i+1,j} - x_{i-1,j})^2 + (y_{i,j+1} - y_{i,j-1})^2}
% \end{equation}

% Taiwan Topology and its Divergence

% QM bias correction with ECDF
    \begin{figure}[tb]
        \centering
            \begin{subfigure}[t]{0.45\columnwidth}
                \centering
                \includegraphics[width=\columnwidth]{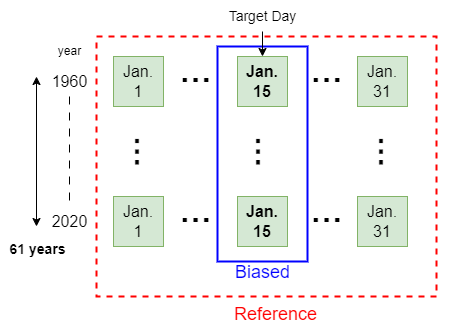}
                \caption{Sampling}
            \end{subfigure}
        \hfill
            \begin{subfigure}[t]{0.45\columnwidth}
                \centering
                \includegraphics[width=\columnwidth]{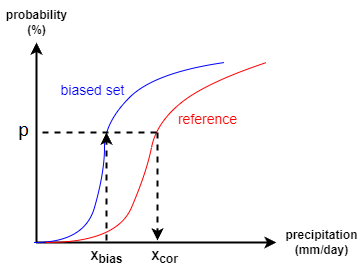}
                \caption{ECDF}
            \end{subfigure}
        \caption{Bias correction with Empirical Quantile Mapping. 
            (a) Sampling of biased and reference data. 
            To construct the ECDFs in (b), the CDF of biased one is obtained based on the target date through 61 years (1960 - 2020) (blue rectangle), and the CDF of reference one is based on whole days within the window through all years (red rectangle). 
            (b) ECDFs of biased and reference data. 
            The blue curve is the CDF of biased data, while the red one is the CDF of reference data. 
            A corrected value $x_{cor}$ is derived from the relationship between the CDF of biased data (bilinearly interpolated) and the CDF of reference one. 
            This process is performed through every grid point of the biased data. 
            Given a biased grid point $x_{bias}$ in target day, the cumulative probability $p$ is obtained according to the CDF of biased data. 
            With the value of $p$, the corrected value $x_{cor}$ is obtained form the inverse function of the CDF of reference data.}
        \label{fig:ECDFandSampling}
    \end{figure}

%%%%%%%%%%%%%%%%%%%%%%%%%%%%%%%%%%%%%%%%%%%%%%%%%% Prediction %%%%%%%%%%%%%%%%%%%%%%%%%%%%%%%%%%%%%%%%%%%%%%%%%%%%%%%%
% Comparison of different models and traditional statistical methods; Scaling Factor = 5
\begin{figure*}[t!]
    \centering
        % Input
        \begin{subfigure}[t]{0.37\columnwidth}
            \centering
            \includegraphics[width=\columnwidth]{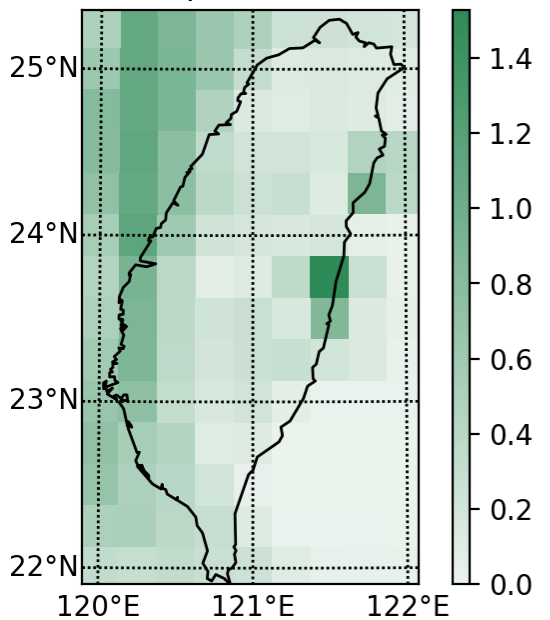}
            \caption{ERA5 Input}
        \end{subfigure}
        % \hfill
        % Ours
        \begin{subfigure}[t]{0.37\columnwidth}
            \centering
            \includegraphics[width=\columnwidth]{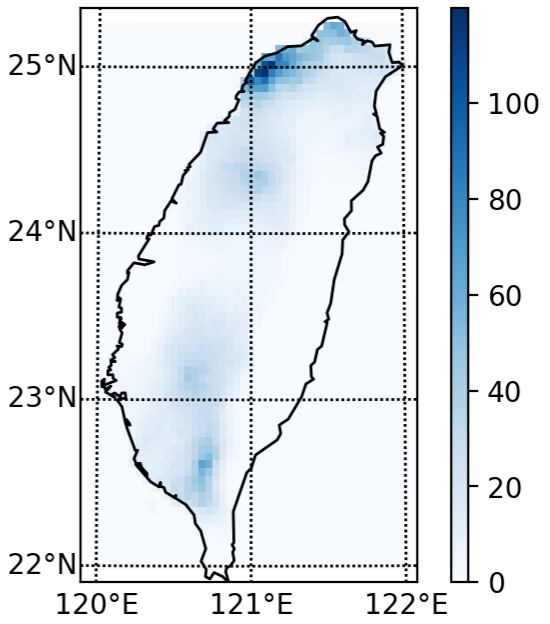}
            \caption{{\textbf{Ours}}}
        \end{subfigure}
        % \hfill
        % \vskip
        % \baselineskip
        % YNet
        \begin{subfigure}[t]{0.37\columnwidth}
            \centering
            \includegraphics[width=\columnwidth]{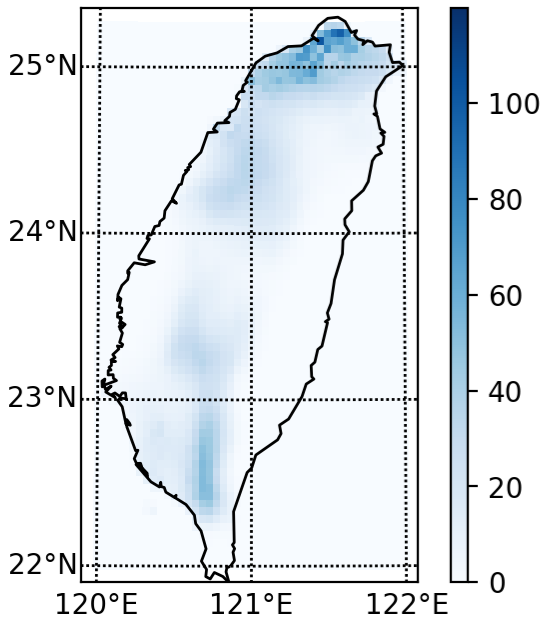}
            \caption{YNet}
        \end{subfigure}
        % \hfill
        % FSRCNN-ESM
        \begin{subfigure}[t]{0.37\columnwidth}
            \centering
            \includegraphics[width=\columnwidth]{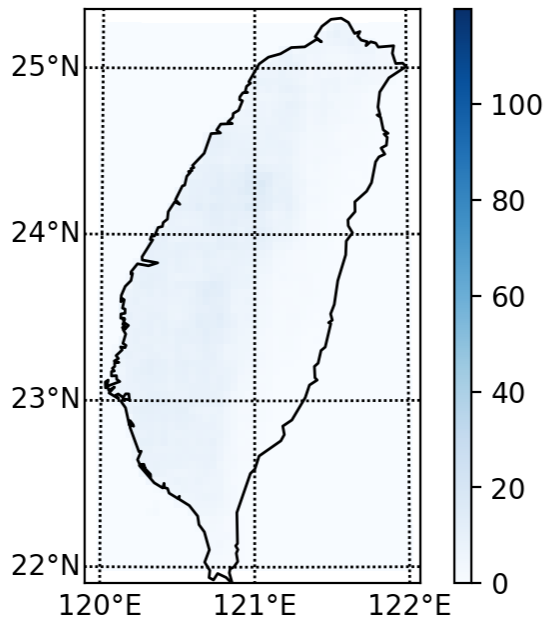}
            \caption{FSRCNN-ESM}
        \end{subfigure}
        \vskip
        \baselineskip
        % Ground Truth
        \begin{subfigure}[t]{0.37\columnwidth}
            \centering
            \includegraphics[width=\columnwidth]{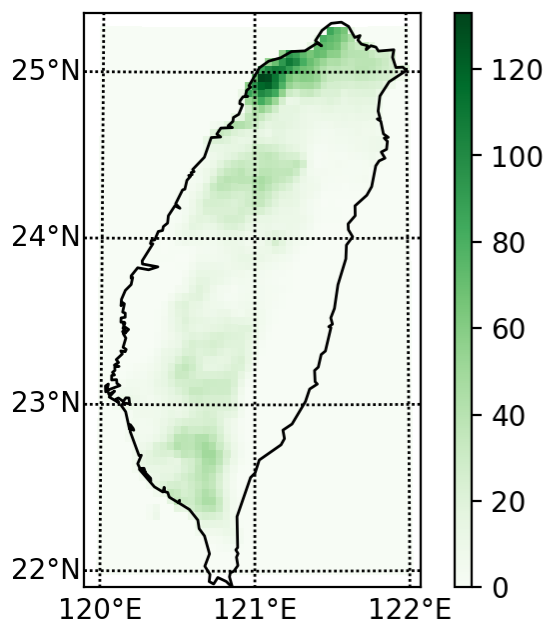}
            \caption{Ground Truth}
        \end{subfigure}
        % \hfill
        % DeepSD
        \begin{subfigure}[t]{0.37\columnwidth}
            \centering
            \includegraphics[width=\columnwidth]{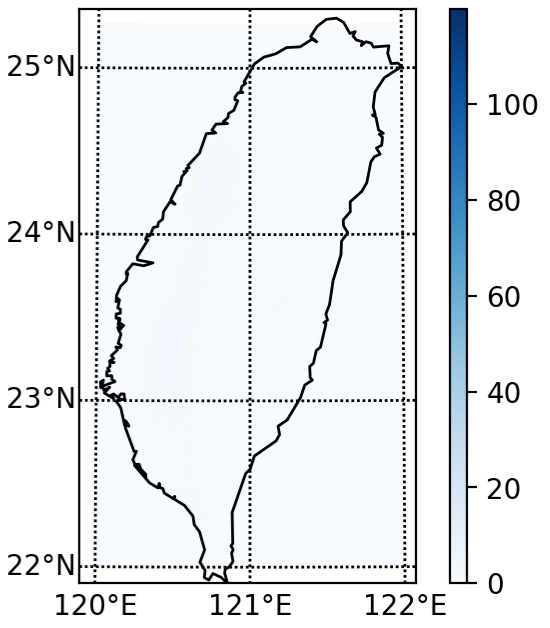}
            \caption{DeepSD}
        \end{subfigure}
        % \hfill
        % \vskip
        % \baselineskip
        % QM
        \begin{subfigure}[t]{0.37\columnwidth}
            \centering
            \includegraphics[width=\columnwidth]{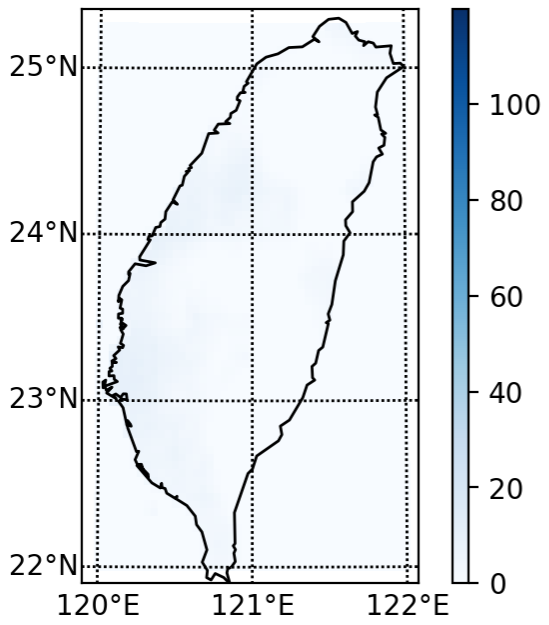}
            \caption{QM}
        \end{subfigure}
        % \hfill
        % BCSD
        \begin{subfigure}[t]{0.37\columnwidth}
            \centering
            \includegraphics[width=\columnwidth]{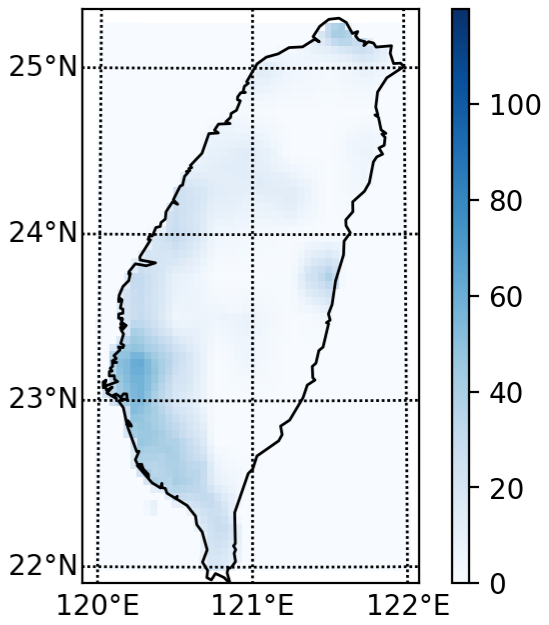}
            \caption{BCSD}
        \end{subfigure}
    \caption{Predictions of an extremely concentrated precipitation case with different climate downscaling methods (scaling factor = 5). 
            (a) The input data of ERA5 reanalysis from ECMWF. 
            (e) The observations from TCCIP, as our groung truth data. 
            Notice that the range of colormap of (a) and (e) are different in order to more clearly show the significant bias problem between the input and correseponding observation. 
            Other subfigures show the super-resolutions with different climate downscaling methods whose color map ranges are aligned with the ground truth's. 
            The deeper blue is, the larger value is. 
            Both (b) and (c) are able to capture the real precipitation range and main spatial distribution in north Taiwan. 
            However, for (d) (f) and (g), the predictions values are pretty low, showing the main challenges in bias correction. 
            As for (h), since it adds the Spatial Disaggregation step with correseponding mean value of observations Eq. (\ref{eq:SD}), the value range are closer to the ground truth, but still fails to predict the main precipitation patterns.}
    \label{fig:diffmodelpredx5}
\end{figure*}

% MAEs of different climate downscaling methods; Scaling Factor = 5
\begin{figure*}[b!]
    \centering
        % Ours
        \begin{subfigure}[t]{0.35\columnwidth}
            \centering
            \includegraphics[width=\columnwidth]{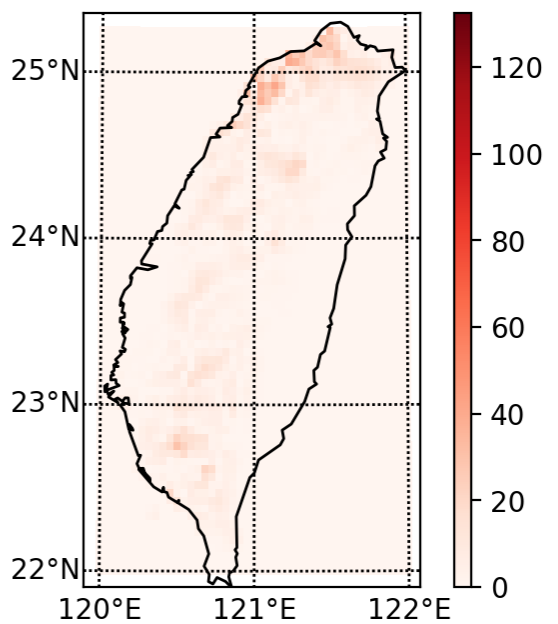}
            \caption{{\textbf{Ours}}}
        \end{subfigure}
        % \hfill
        % YNet
        \begin{subfigure}[t]{0.35\columnwidth}
            \centering
            \includegraphics[width=\columnwidth]{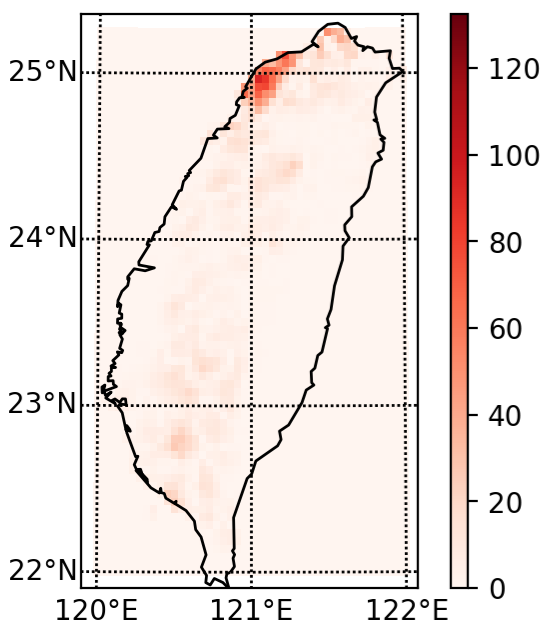}
            \caption{YNet}
        \end{subfigure}
        % \hfill
        % FSRCNN-ESM
        \begin{subfigure}[t]{0.35\columnwidth}
            \centering
            \includegraphics[width=\columnwidth]{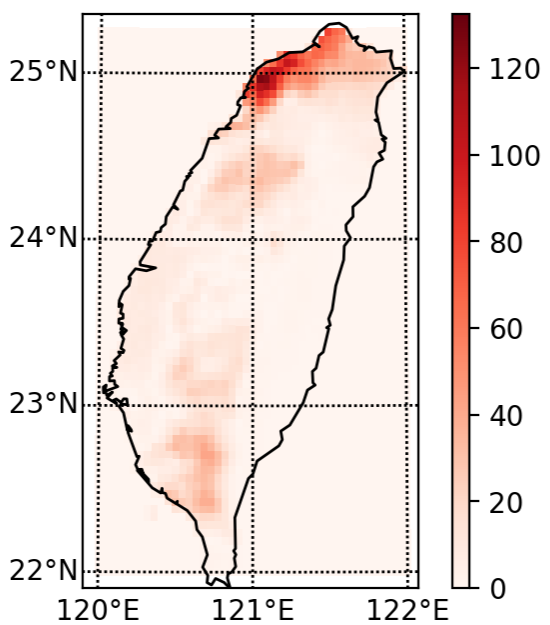}
            \caption{FSRCNN-ESM}
        \end{subfigure}
        \vskip
        \baselineskip
        % DeepSD
        \begin{subfigure}[t]{0.35\columnwidth}
            \centering
            \includegraphics[width=\columnwidth]{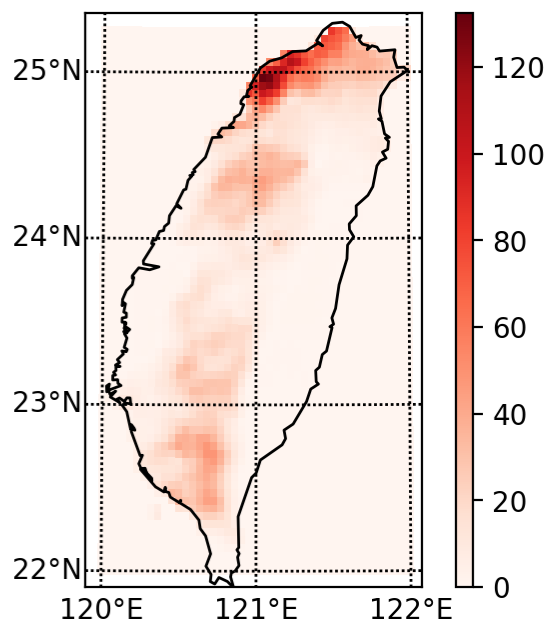}
            \caption{DeepSD}
        \end{subfigure}
        % \hfill
        % QM
        \begin{subfigure}[t]{0.35\columnwidth}
            \centering
            \includegraphics[width=\columnwidth]{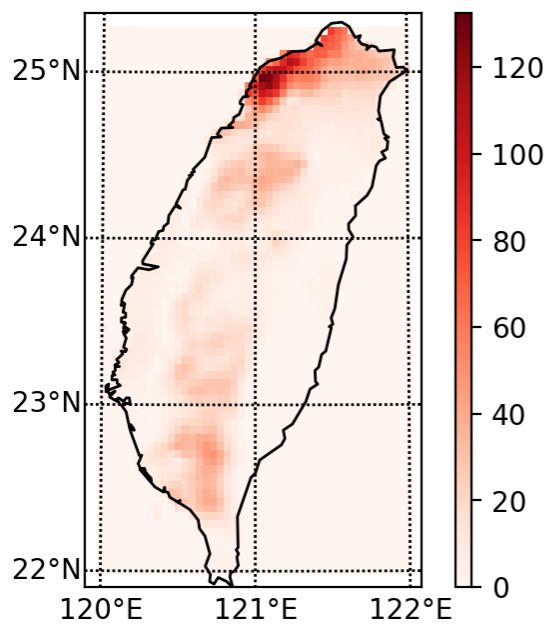}
            \caption{QM}
        \end{subfigure}
        % \hfill
        % BCSD
        \begin{subfigure}[t]{0.35\columnwidth}
            \centering
            \includegraphics[width=\columnwidth]{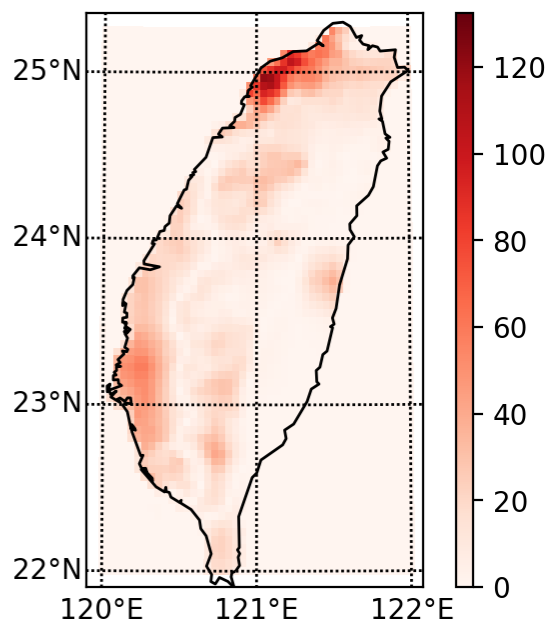}
            \caption{BCSD}
        \end{subfigure}
    \caption{MAEs of an extremely concentrated precipitation case with different climate downscaling methods (scaling factor = 5). 
            Continued with Fig. \ref{fig:diffmodelpredx5}, the subfigures show the differences (MAEs) between ground truth and SR results of different climate downscaling methods. 
            The deeper red is, the larger difference is, and the more similar to the image pattern of ground truth is, the worse performance it is. 
            The ground truth precipitation values are distributed in the west of Taiwan, and specifically, higher values in north-west part.
            Ours shows the lowest MAE among all predictions.}
    \label{fig:diffmodelx5mae}
\end{figure*}

\section{Experiment}
\label{sec:exp}
    \subsection{Setup}
        \subsubsection{Model Implementation}
        We implemented our model in the Tensorflow framework.
        There are a total of 22,281 daily precipitation data and are randomly separated by 80\% for training, 10\% for validation, and the other 10\% for testing. 
        The batch size is set to be 64 and an epoch is limited to a maximum training loop by 1,000 with an early-stop setting by patience equal to 60. 
        We use the mean square error (MSE) as our loss function Eq. (\ref{eq:lossFunction}). 
        The model training is optimized by Adam Optimizer with default settings and a learning rate of $10^{-4}$. 
        The training process was performed on NVIDIA® Tesla V100 GPU.
        \begin{equation}
            \label{eq:lossFunction}
                MSE = \frac{1}{H\times W}\sum_{i=1}^{H}\sum_{j=1}^{W}\|\hat{Y}_{i,j}-Y_{i,j}\|^2
        \end{equation}
        where $H$ and $W$ are the height and width of the prediction, and $\hat{Y}_{i,j}$ is the corresponding ground truth of the prediction $Y_{i,j}$.

        \subsubsection{Metrics}
        Peak Signal to Noise Ratio (PSNR) is a commonly used metric in the image field. 
        But in scientific data, since there is no theoretical maximum value, PSNR may not meet our needs. 
        Hence, we use mean absolute error (MAE), root mean square error (RMSE), Pearson Correlation (Corr.), and structural similarity index (SSIM) proposed by \cite{SSIM} as our metrics. 
        Lower MAE and RMSE indicate better accuracy between the predictions and observations, while higher Pearson Correlation and SSIM indicate stronger relationships between the predictions and observations. 
        
        Also, we evaluate the results with forecast indicators: Probability of Detection (POD), False Alarm Ratio (FAR) and Threat Score (TS) (e.g., \cite{qof01, qof02}). 
        The threshold is set as 0.1 mm for calculating the forecast indicators, as a critical value of rainfall existence, to calculate indicators. 
        A higher POD indicates a higher percentage of correctly predicted events, and a higher TS indicates a higher skill in capturing both hits and correct negatives, while minimizing false alarms and misses.
        A lower FAR indicates a lower percentage of false alarms or incorrect forecasts.

        \subsubsection{Alternative Approaches}
        We pick two statistical climate downscaling methods: QM by \cite{QM} and BCSD by \cite{BCSD}. Additionally,  we re-train other three deep learning based models, DeepSD, FSRCNN-ESM, and YNet for comparison. 
        We briefly introduce the QM and BCSD algorithm below:  
        let $x^{bias}_{i,j}$ be the biased value of precipitation at a grid point $(i,j)$ in the $k^{th}$ day. 
        And let $F^{bias}_{i,j}$ be the CDF of the grid point in biased dataset, and $F^{ref}_{i,j}$ of reference, correspondingly. 
        To get the corrected value $x^{cor}_{i,j}$ take the value in reference dataset of the same CDF value as corrected one, that is:
            \begin{equation}
                \label{eq:QM}
                x^{cor}_{i,j} = InvF^{ref}_{i,j} [F^{bias}_{i,j} (x^{bias}_{i,j}) ]\qquad \forall \;x^{bias}_{i,j} \in X^{bias}_{k}
            \end{equation}
        where $InvF(\cdot)$ is the inverse of CDF, $X^{bias}_{k}$ is the set of biased grid points in the $k^{th}$ day, as shown in Fig. \ref{fig:ECDFandSampling}b. 
        In general, the correction is based on the resolution of the biased dataset, and in this case the reference dataset is degraded to 25 km, the same as the biased one. 
        Also, the CDF of the target day is calculated on each grid point over the years within a time window. 
        The window size is set to be 15 as \cite{TCCIPgrid} suggested, that is, before and after the $k^{th}$ day, $k\pm15$. 
        We then take all the data within the window over years as to-be-corrected one, and corresponding days over reference one to get $F^{ref}$, as shown in Fig. \ref{fig:ECDFandSampling}a. 
        % For example, if we would like to corrected the precipitation data at Jan. $16^{th}$, we will take all Jan. $16^{th}$ rainfall data from 1960 to 2020 as the biased data. 
        % On the other hand, the reference data would be Jan. data from 1960 to 2020, shown in \cref{fig:ECDFandSampling} (a). 
        % As for the date like Jan $1^{st}$, where half of the range of bias correction window has no historical data, we treat only the other half within the window, that is, from Jan. $1^{st}$ to Jan. $16^{th}$, as the references for bias correction. 
        
        BCSD algorithm further adds a Spatial Disaggregation (SD) step after QM shown in Eq. (\ref{eq:SD}). 
        SD step performs the interpolation of the difference of target daily rainfall $X^{cor}_{k}$ and the average of degraded reference $Y_{l}$. 
        Then, multiply with the SD factor, which is the ratio between the means of original reference rainfall $Y_{h}$ and the degraded one $Y_{l}$. 
        Eventually, add original reference to the product mentioned above to get the final value $Z_{k}$, which is:
        \begin{equation}
            \label{eq:SD}
            Z_{k} = Y_{h} + Intp(X^{cor}_{k} - Y_{l}) * \frac{Y_{h}}{Y_{l} + 1}
        \end{equation}
        where $Intp(\cdot)$ is the interpolation method, and $\frac{Y_{h}}{Y_{l} + 1}$ is the scaling factor which is modified by plus one to avoid unrealistic value of dry days.

    \subsection{Results}
    The predictions of various climate downscaling methods with different scaling factors are shown in Table \ref{tb:diffmodels}. 
    Our model performs better in the metrics and indicators, except the POD in each scaling factor. 
    We notice that BCSD has the highest POD values, which is \textbf{1.0000}. 
    It is because POD only considers the number of points that predictions and corresponding ground truths are both greater than the threshold (0.1 mm). 
    To an extreme case, if the prediction is all ones, the POD would also be \textbf{1.0000}. 
    This overestimation might result from the multiplication in Eq. (\ref{eq:SD}). 
    And we can tell such errors from the other two indicators, FAR and TS. 
    As for DeepSD and FSRCNNESM, after we re-train them with our dataset, they results in lower performance. 
    Since YNet was initially designed and evaluated in continental precipitation ensemble data, for single channel and non-continental data, ours shows better performance in such scenario. 

    Among the results, we pick three representatives of precipitation distribution in Taiwan, ordered by the concentration of precipitation. 
    Cooperated with the domain experts, the model with the scaling factor of 5 is looked for in current climatological research, since the resolution of the first-hand observation data is the closest to this factor. 
    Therefore, we pick this scaling factor for showcases. 

\begin{figure*}[h!]
    \centering
        % Input
        \begin{subfigure}[t]{0.37\columnwidth}
            \centering
            \includegraphics[width=\columnwidth]{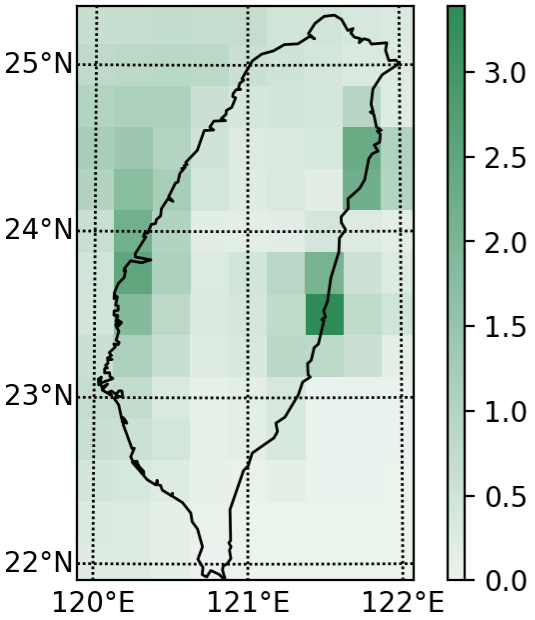}
            \caption{ERA5 Input}
        \end{subfigure}
        % /hfill
        % Ours
        \begin{subfigure}[t]{0.37\columnwidth}
            \centering
            \includegraphics[width=\columnwidth]{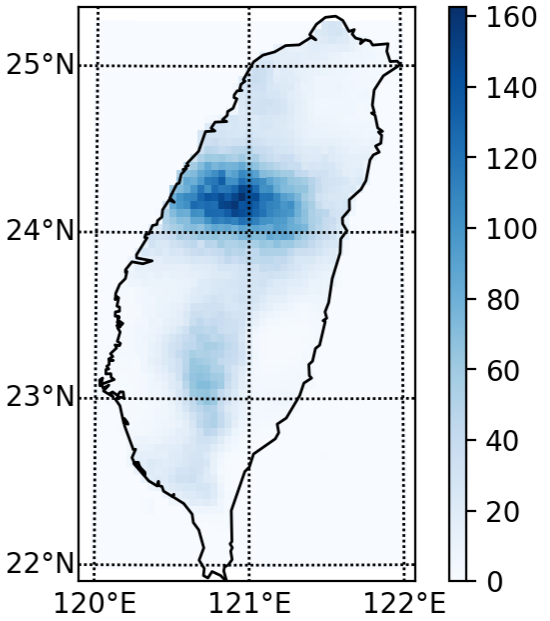}
            \caption{{\textbf{Ours}}}
        \end{subfigure}
        % /hfill
        % YNet
        \begin{subfigure}[t]{0.37\columnwidth}
            \centering
            \includegraphics[width=\columnwidth]{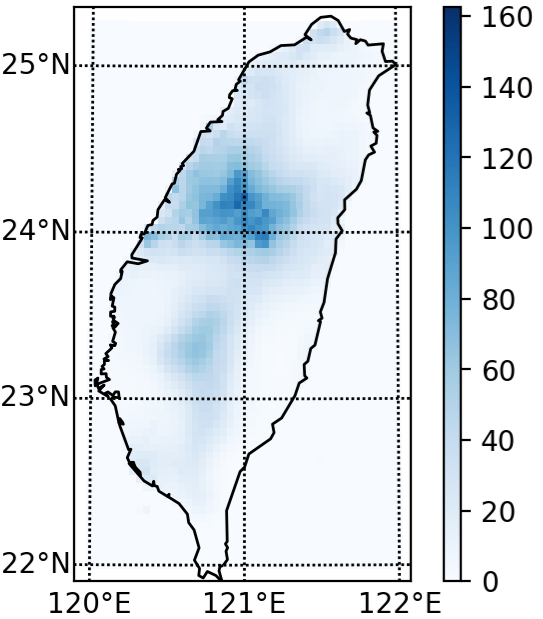}
            \caption{YNet}
        \end{subfigure}
        % /hfill
        % FSRCNN-ESM
        \begin{subfigure}[t]{0.37\columnwidth}
            \centering
            \includegraphics[width=\columnwidth]{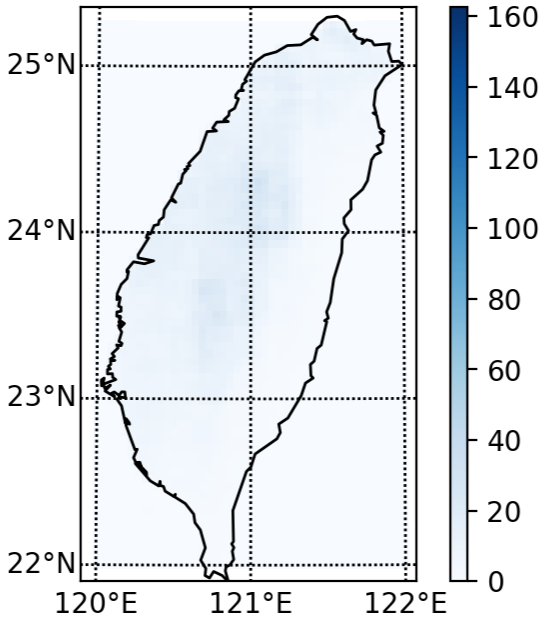}
            \caption{FSRCNN-ESM}
        \end{subfigure}
        \vskip
        \baselineskip
        % Ground Truth
        \begin{subfigure}[t]{0.37\columnwidth}
            \centering
            \includegraphics[width=\columnwidth]{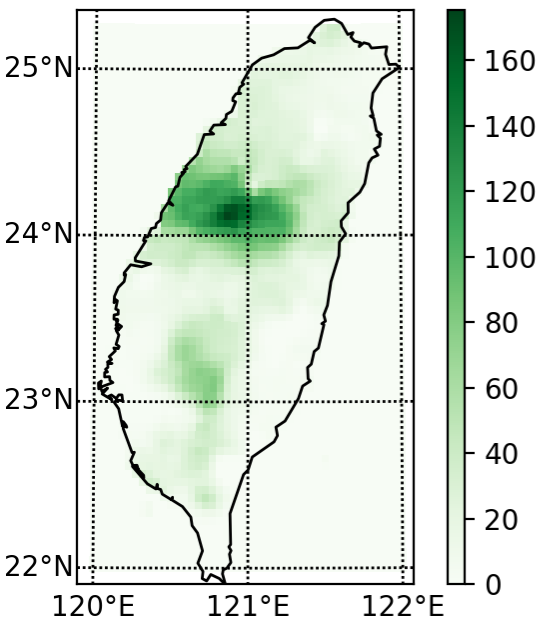}
            \caption{Ground Truth}
        \end{subfigure}
        % /hfill
        % DeepSD
        \begin{subfigure}[t]{0.37\columnwidth}
            \centering
            \includegraphics[width=\columnwidth]{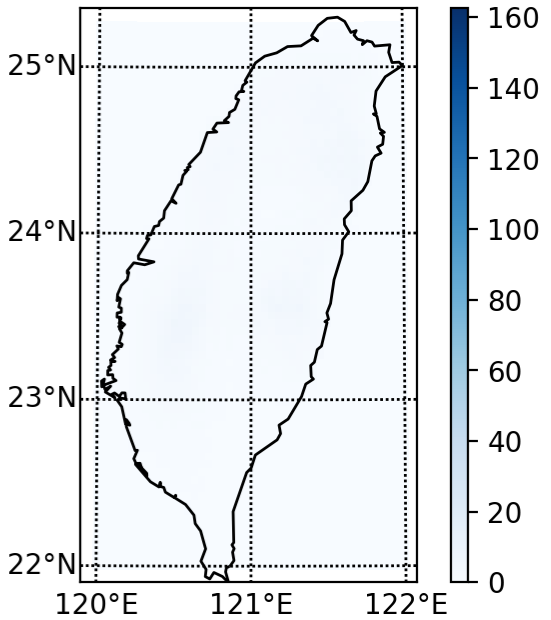}
            \caption{DeepSD}
        \end{subfigure}
        % /hfill
        % QM
        \begin{subfigure}[t]{0.37\columnwidth}
            \centering
            \includegraphics[width=\columnwidth]{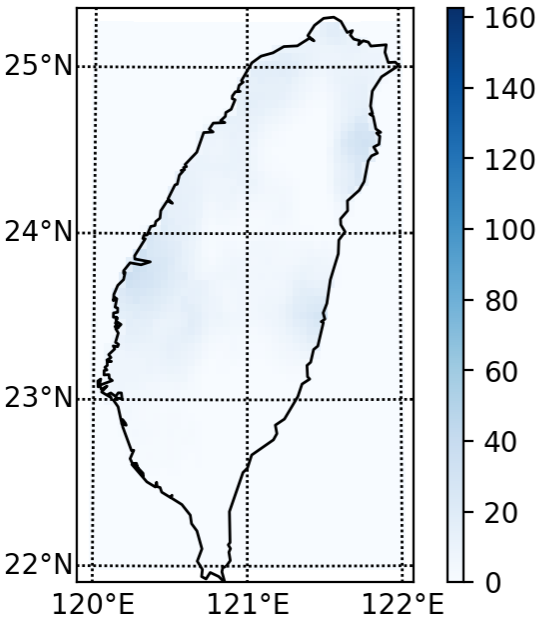}
            \caption{QM}
        \end{subfigure}
        % \hfill
        % BCSD
        \begin{subfigure}[t]{0.37\columnwidth}
            \centering
            \includegraphics[width=\columnwidth]{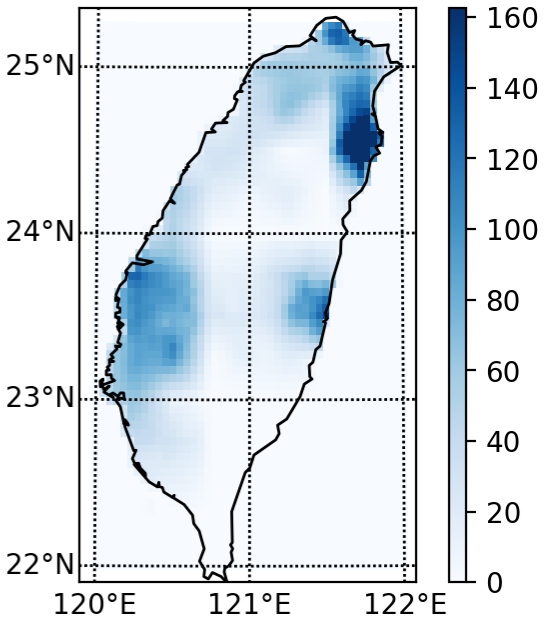}
            \caption{BCSD}
        \end{subfigure}
    \caption{Predictions of an uneven distributed precipitation case with different climate downscaling methods (scaling factor = 5). 
            (a) The input data of ERA5 reanalysis from ECMWF. 
            (e) The observations from TCCIP, as our groung truth data.}
    \label{fig:diffmodelpredx5_uneven_rain}

% MAEs example 2

    \centering
        % Ours
        \begin{subfigure}[t]{0.35\columnwidth}
            \centering
            \includegraphics[width=\columnwidth]{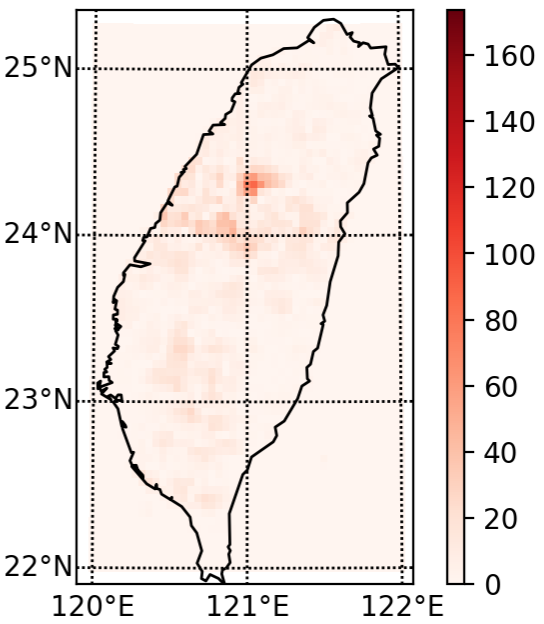}
            \caption{{\textbf{Ours}}}
        \end{subfigure}
        % \hfill
        % YNet
        \begin{subfigure}[t]{0.35\columnwidth}
            \centering
            \includegraphics[width=\columnwidth]{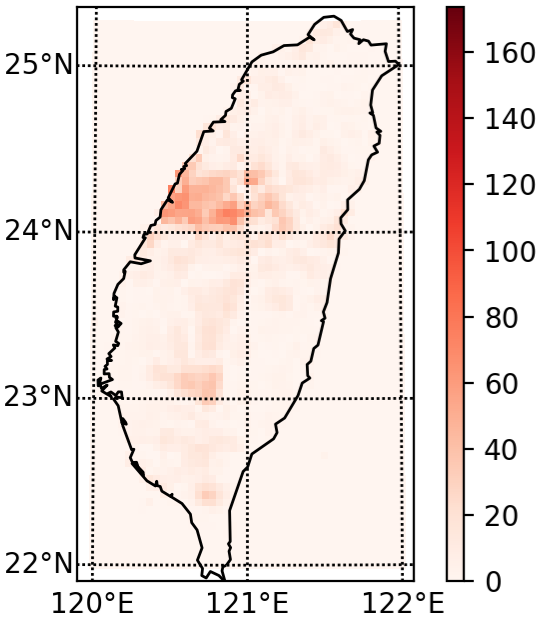}
            \caption{YNet}
        \end{subfigure}
        % \hfill
        % FSRCNN-ESM
        \begin{subfigure}[t]{0.35\columnwidth}
            \centering
            \includegraphics[width=\columnwidth]{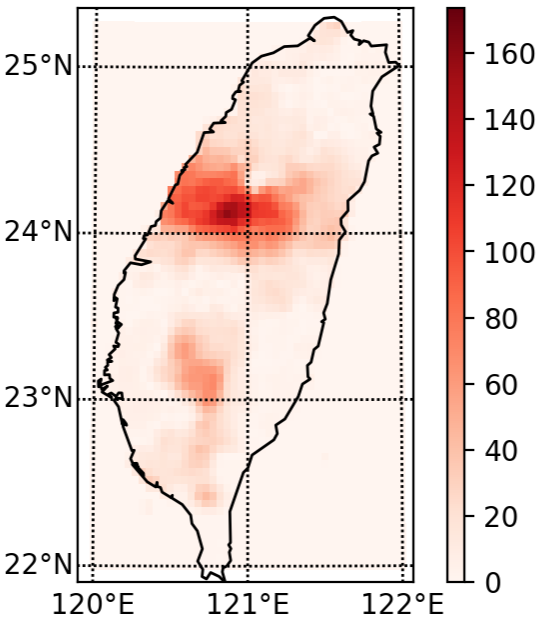}
            \caption{FSRCNN-ESM}
        \end{subfigure}
        \vskip
        \baselineskip
        % DeepSD
        \begin{subfigure}[t]{0.35\columnwidth}
            \centering
            \includegraphics[width=\columnwidth]{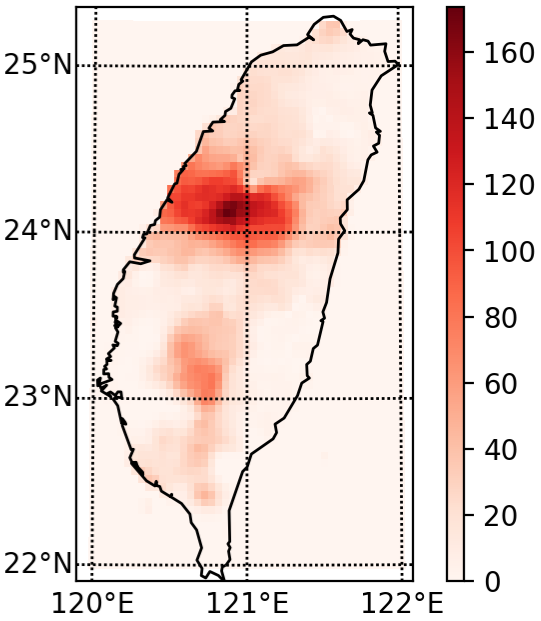}
            \caption{DeepSD}
        \end{subfigure}
        % \hfill
        % QM
        \begin{subfigure}[t]{0.35\columnwidth}
            \centering
            \includegraphics[width=\columnwidth]{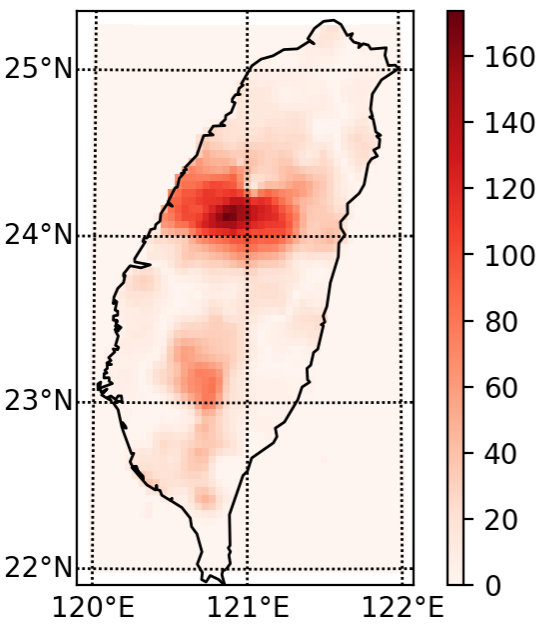}
            \caption{QM}
        \end{subfigure}
        % \hfill
        % BCSD
        \begin{subfigure}[t]{0.35\columnwidth}
            \centering
            \includegraphics[width=\columnwidth]{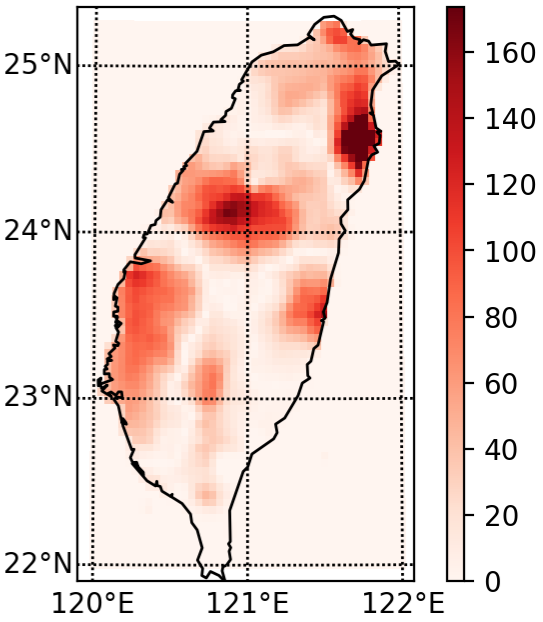}
            \caption{BCSD}
        \end{subfigure}
    \caption{MAEs of an uneven distributed precipitation case with different climate downscaling methods (scaling factor = 5).Continued with Fig. \ref{fig:diffmodelpredx5_uneven_rain}, the subfigures show the differences (MAEs) between ground truth and SR results of different climate downscaling methods. The deeper red is, the larger difference is, and the more similar to the image pattern of ground truth is, the worse performance it is.}
    \label{fig:diffmodelx5mae_uneven_rain}
\end{figure*}

\begin{figure*}
    \centering
        % Input
        \begin{subfigure}[t]{0.37\columnwidth}
            \centering
            \includegraphics[width=\columnwidth]{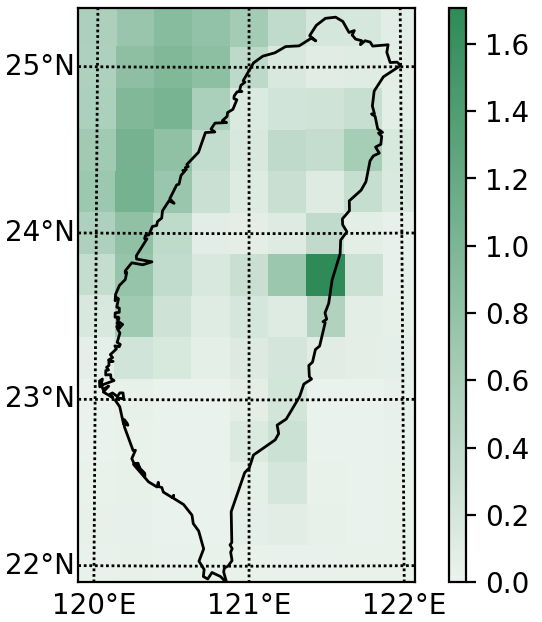}
            \caption{ERA5 Input}
        \end{subfigure}
        % /hfill
        % Ours
        \begin{subfigure}[t]{0.37\columnwidth}
            \centering
            \includegraphics[width=\columnwidth]{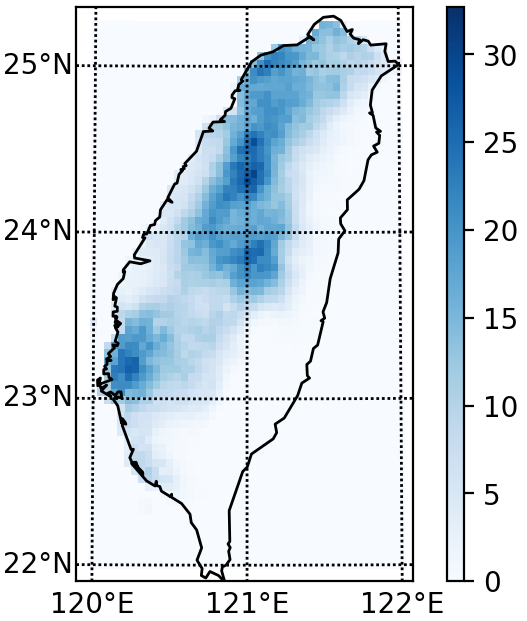}
            \caption{{\textbf{Ours}}}
        \end{subfigure}
        % /hfill
        % YNet
        \begin{subfigure}[t]{0.37\columnwidth}
            \centering
            \includegraphics[width=\columnwidth]{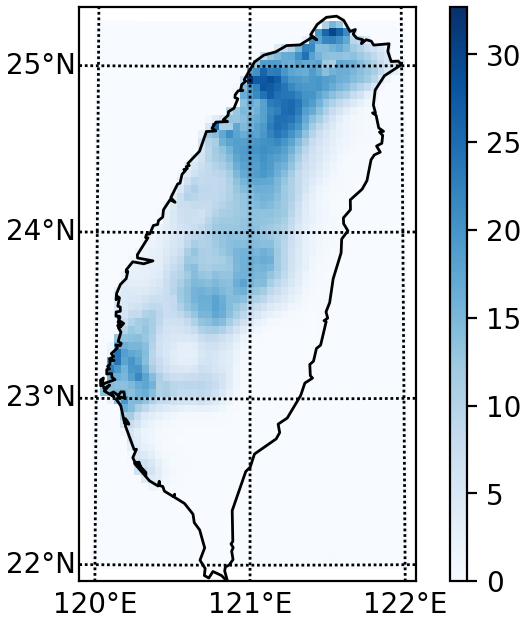}
            \caption{YNet}
        \end{subfigure}
        % /hfill
        % FSRCNN-ESM
        \begin{subfigure}[t]{0.37\columnwidth}
            \centering
            \includegraphics[width=\columnwidth]{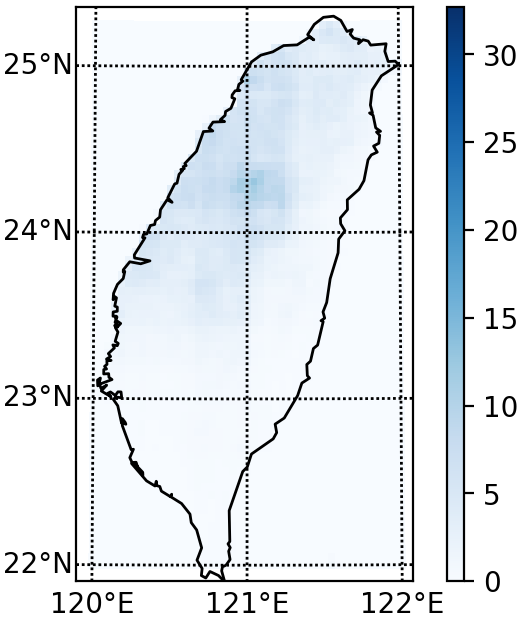}
            \caption{FSRCNN-ESM}
        \end{subfigure}
        \vskip
        \baselineskip
        % Ground Truth
        \begin{subfigure}[t]{0.37\columnwidth}
            \centering
            \includegraphics[width=\columnwidth]{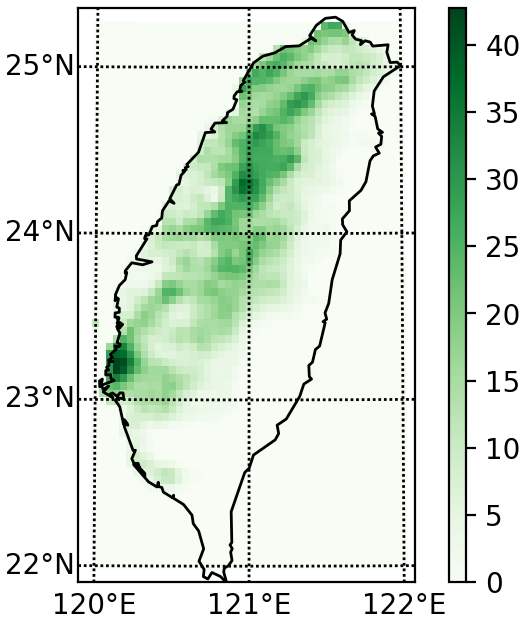}
            \caption{Ground Truth}
        \end{subfigure}
        % /hfill
        % DeepSD
        \begin{subfigure}[t]{0.37\columnwidth}
            \centering
            \includegraphics[width=\columnwidth]{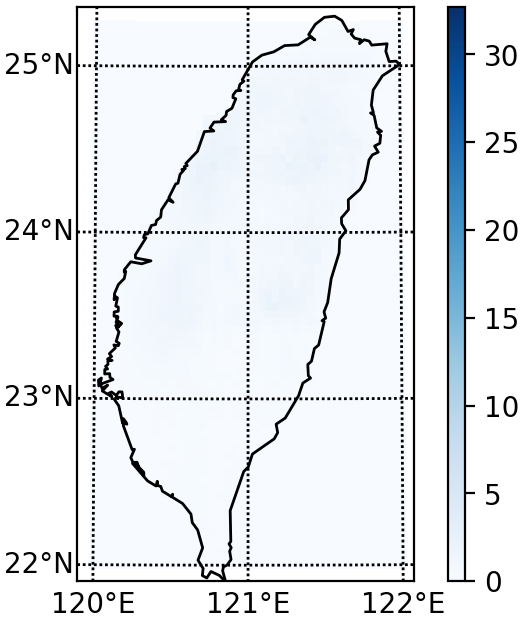}
            \caption{DeepSD}
        \end{subfigure}
        % /hfill
        % QM
        \begin{subfigure}[t]{0.37\columnwidth}
            \centering
            \includegraphics[width=\columnwidth]{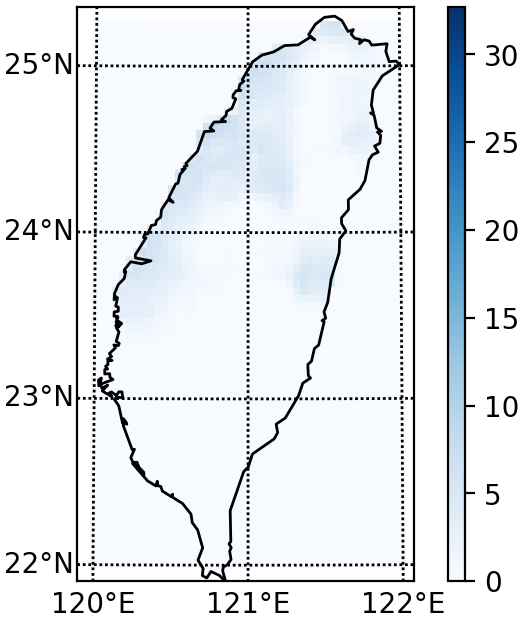}
            \caption{QM}
        \end{subfigure}
        % \hfill
        % BCSD
        \begin{subfigure}[t]{0.37\columnwidth}
            \centering
            \includegraphics[width=\columnwidth]{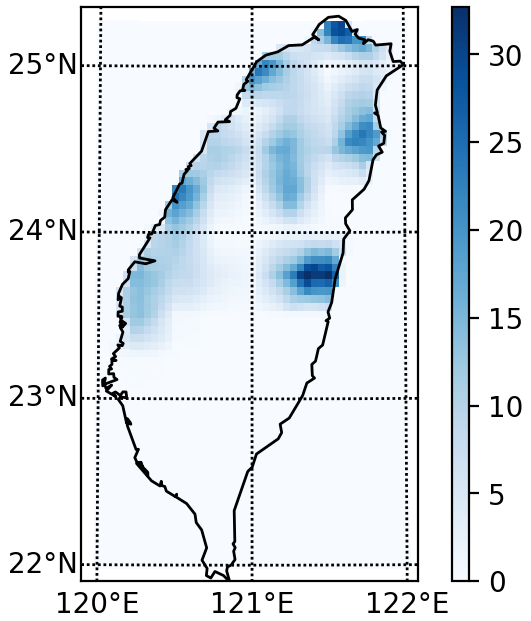}
            \caption{BCSD}
        \end{subfigure}
    \caption{Predictions of a relatively even distributed precipitation case with different climate downscaling methods (scaling factor = 5). 
            (a) The input data of ERA5 reanalysis from ECMWF. 
            (e) The observations from TCCIP, as our groung truth data.}
    \label{fig:diffmodelpredx5_even_rain}
\end{figure*}

\begin{figure*}[tb]
    \centering
        % Ours
        \begin{subfigure}[t]{0.35\columnwidth}
            \centering
            \includegraphics[width=\columnwidth]{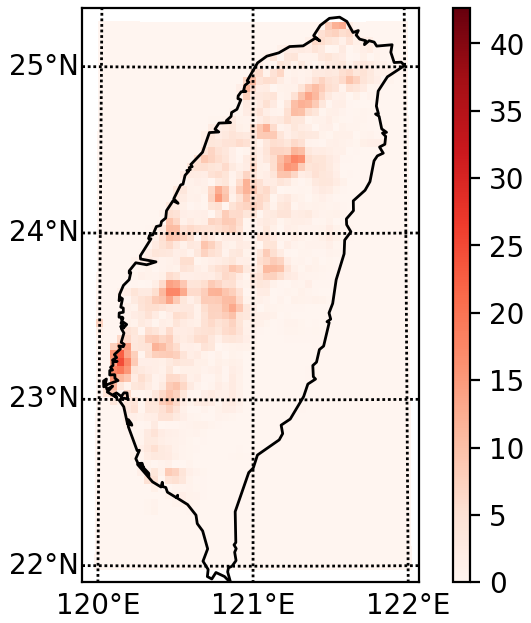}
            \caption{{\textbf{Ours}}}
        \end{subfigure}
        % \hfill
        % YNet
        \begin{subfigure}[t]{0.35\columnwidth}
            \centering
            \includegraphics[width=\columnwidth]{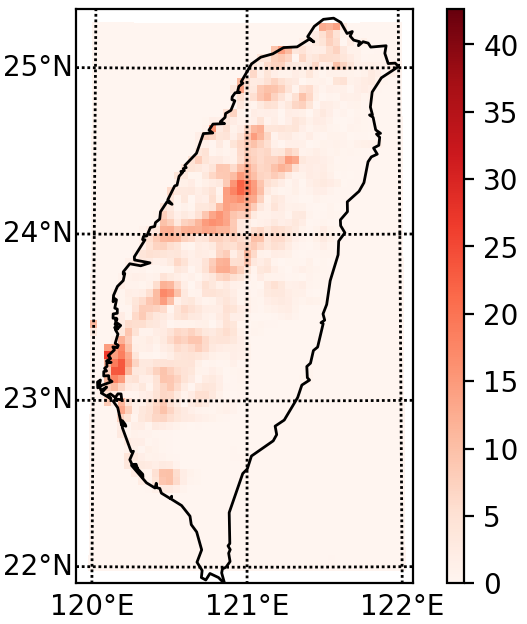}
            \caption{YNet}
        \end{subfigure}
        % \hfill
        % FSRCNN-ESM
        \begin{subfigure}[t]{0.35\columnwidth}
            \centering
            \includegraphics[width=\columnwidth]{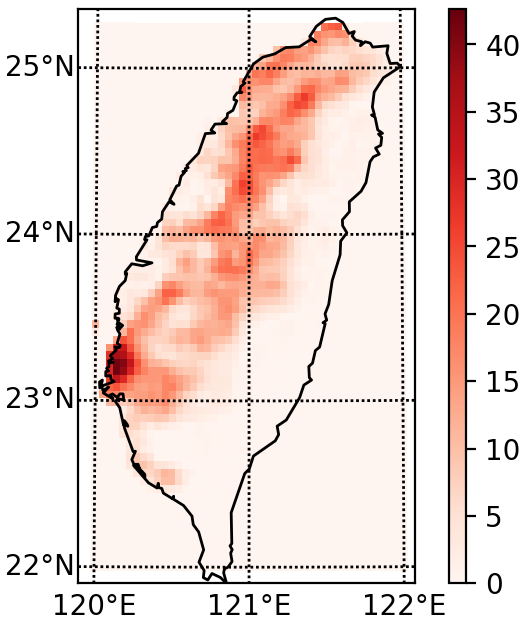}
            \caption{FSRCNN-ESM}
        \end{subfigure}
        \vskip
        \baselineskip
        % DeepSD
        \begin{subfigure}[t]{0.35\columnwidth}
            \centering
            \includegraphics[width=\columnwidth]{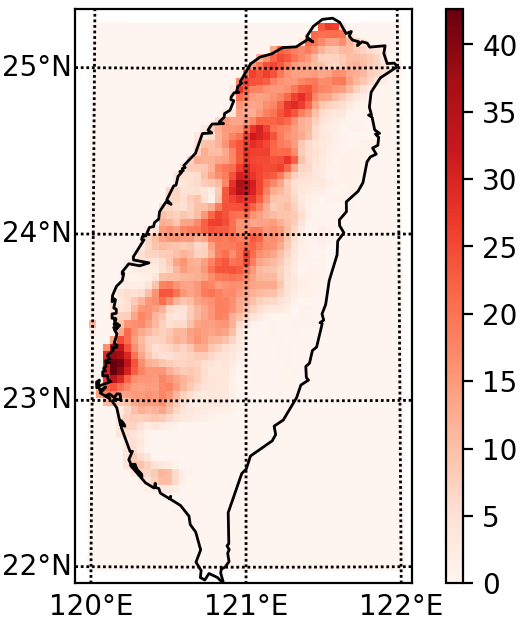}
            \caption{DeepSD}
        \end{subfigure}
        % \hfill
        % QM
        \begin{subfigure}[t]{0.35\columnwidth}
            \centering
            \includegraphics[width=\columnwidth]{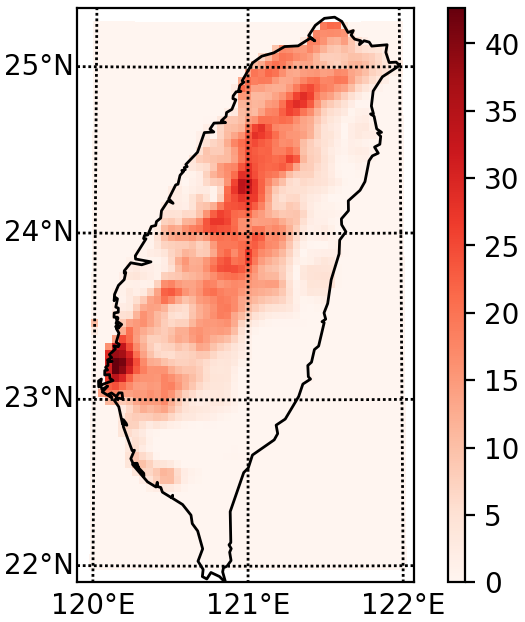}
            \caption{QM}
        \end{subfigure}
        % \hfill
        % BCSD
        \begin{subfigure}[t]{0.35\columnwidth}
            \centering
            \includegraphics[width=\columnwidth]{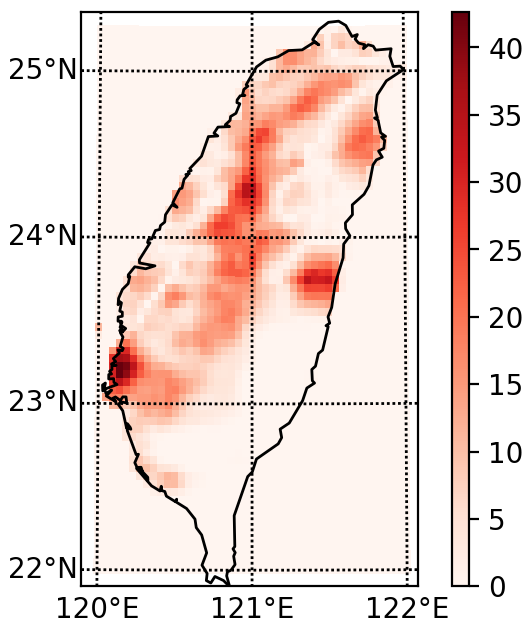}
            \caption{BCSD}
        \end{subfigure}
    \caption{MAEs of a relatively even distributed precipitation case (scaling factor = 5). Continued with Fig. \ref{fig:diffmodelpredx5_even_rain}, the subfigures show the differences (MAEs) between ground truth and SR results of different climate downscaling methods. The deeper red is, the larger difference is, and the more similar to the image pattern of ground truth is, the worse performance it is.}
    \label{fig:diffmodelx5mae_even_rain}
\end{figure*}

    Fig. \ref{fig:diffmodelpredx5} shows the predictions of extremely concentrated rainy day in Jun. 14, 2016. 
    The ERA5 reanalysis data in Fig. \ref{fig:diffmodelpredx5}a infers a concentrated rain at Hualian city (23.8$^{\circ}$N, 121.4$^{\circ}$E), while the corresponding ground truth in Fig. \ref{fig:diffmodelpredx5}e concentrates at Taoyuan city (25$^{\circ}$N, 121$^{\circ}$E). 
    For other cities, there are scattered rainfall in the west, but it merely rains in the east. 
    Our model is able to do the correction of concentrated rainy spots and adjust to realistic precipitation values (Fig. \ref{fig:diffmodelpredx5}b). 
    Although YNet also corrects the concentrated rainy locations, the precipitation values are lower predicted (Fig. \ref{fig:diffmodelpredx5}c). 
    While others like FSRCNN-ESM and DeepSD predicts much lower precipitation values that they shows nearly zeros covering the whole island, and so do QM and BCSD methods. 
    Fig. \ref{fig:diffmodelx5mae} clearly shows the differences between realistic precipitation values and predictions. 
    The second example in Fig. \ref{fig:diffmodelpredx5_uneven_rain} shows the predictions of a heavy rain over the entire Taichung city (the central part of Taiwan) in May 19, 2019. 
    Similarly, our model has predicted correctly for the main rainy regions, and has corrected to realistic precipitation values. 
    Although YNet also corrects the concentrated rainy locations, the precipitation area are smaller predicted (Fig. \ref{fig:diffmodelpredx5_uneven_rain}c). 
    While others are unable to correct to realist values, except BCSD which has multiplied with the scaling factor between observations and biased data in Eq. (\ref{eq:SD}) but the rainfall location is not correctly predicted. 
    The last example in Fig. \ref{fig:diffmodelpredx5_even_rain} shows the predictions of a relatively even-distributed light rains over north and east part of Taiwan in Apr. 18, 2016. 
    The rainy regions are pretty scattered and discrete precipitation makes the errors of predictions be shown in a spotty pattern (Fig. \ref{fig:diffmodelx5mae_even_rain}). 
    Ours and YNet are giving an east rainfall distribution correctly but are failed to precisely capture the discrete patterns. 
    Since the loss function is set as MSE, it tends to generate a smoother result. 
    In other words, it might be difficult for models to precisely predict a discrete, scattered pattern.

    \subsection{Parameter Study}
    To find optimal model structure, we did the following parameter studies below, including model size, type of upscaling layer and showed significant influences of the topography:
               
        \subsubsection{Model Size: Number of Layers}
\begin{figure}[t!]
    \centering
    % training loss
    \begin{subfigure}[t]{0.475\columnwidth}
        \centering
        \includegraphics[width=\columnwidth]{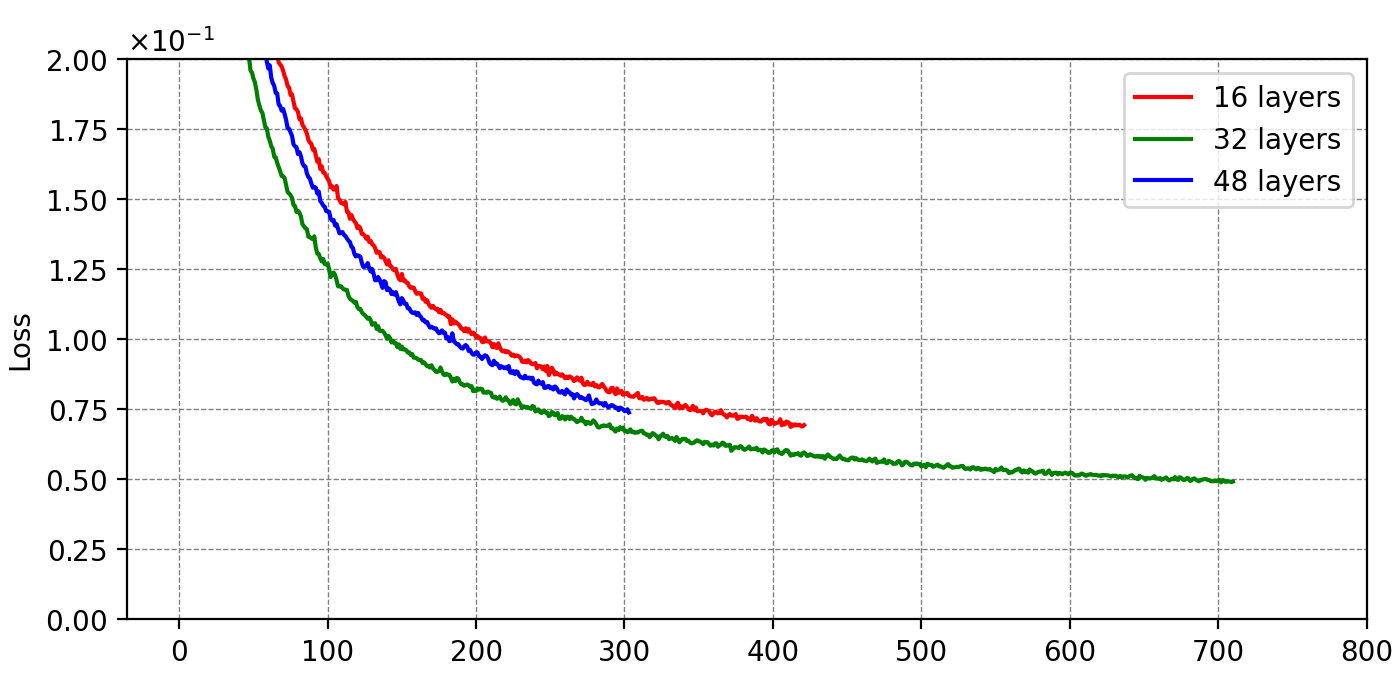}
        \caption{Training Loss}
    \end{subfigure}
    % Validation Loss
    \begin{subfigure}[t]{0.475\columnwidth}
        \centering
        \includegraphics[width=\columnwidth]{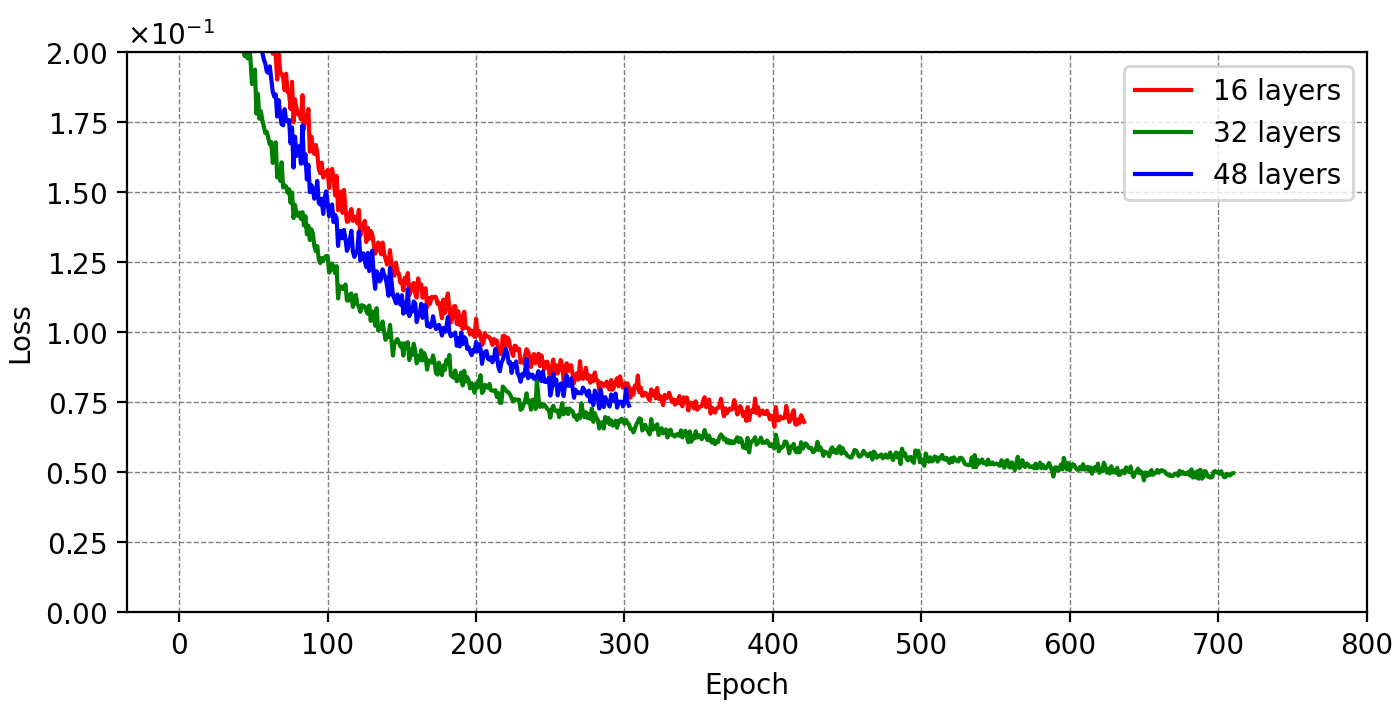}
        \caption{Validation Loss}
    \end{subfigure}
    \caption{Losses of different number of layers. 
            The training hyperparameters are set as epochs = 1,000, batch size = 64, and with an early stopping = 60. The 48-layer one has higher losses than the 32-layer one, which reflects the difficulties in optimization. 
            While the 16-layer one has the highest losses, this indicates the lack of capability.}
    \label{fig:diffnsloss}
\end{figure}

        As a general insight of machine learning, model size or the number of hyperparameters is highly correlated to output quality. 
        Smaller model might end up with higher losses and poor capability, while larger model has more flexibility, but may cause overfitting. 
        We can tell if the model size is proper for the scenario through comparisons of the training and validation losses of different model sizes. 
        We tested three different numbers of convolutional layers in our model spine, that is 16, 32, and 48, to find the most suitable size for our configuration. 
        The training and validation losses are shown in Fig. \ref{fig:diffnsloss}. 
        The 48-layer one stops the earliest, but has higher losses than the 32-layer one, which may inform the difficulties in optimization. 
        The highest losses occur in the 16-layer one which indicates its lack of capability among others in handling the climate downscaling of heterogeneous data. 

\begin{figure}[t!]
    \centering
        % MAE
        \begin{subfigure}[t]{0.24\columnwidth}
            \centering
            \includegraphics[width=\columnwidth]{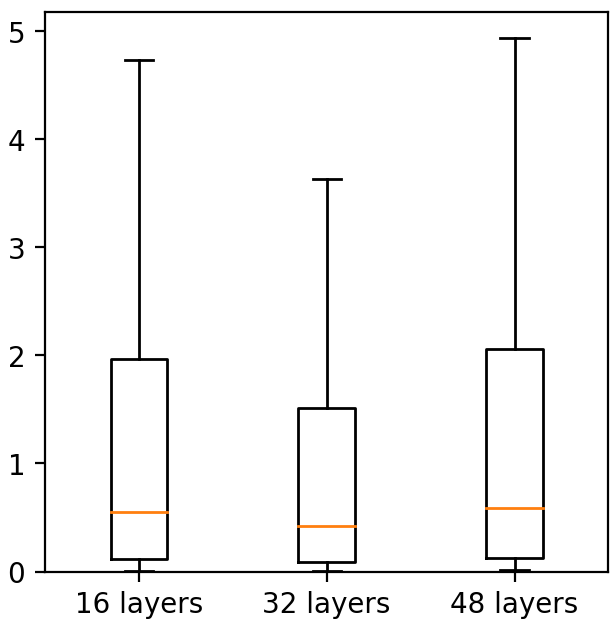}
            \caption{MAE}
        \end{subfigure}
        % RMSE
        \begin{subfigure}[t]{0.24\columnwidth}
            \centering
            \includegraphics[width=\columnwidth]{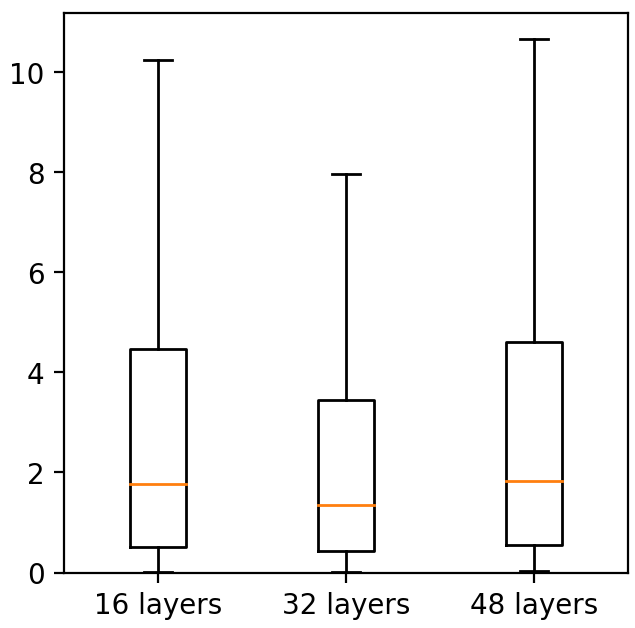}
            \caption{RMSE}
        \end{subfigure}
        % CORR
        \begin{subfigure}[t]{0.24\columnwidth}
            \centering
            \includegraphics[width=\columnwidth]{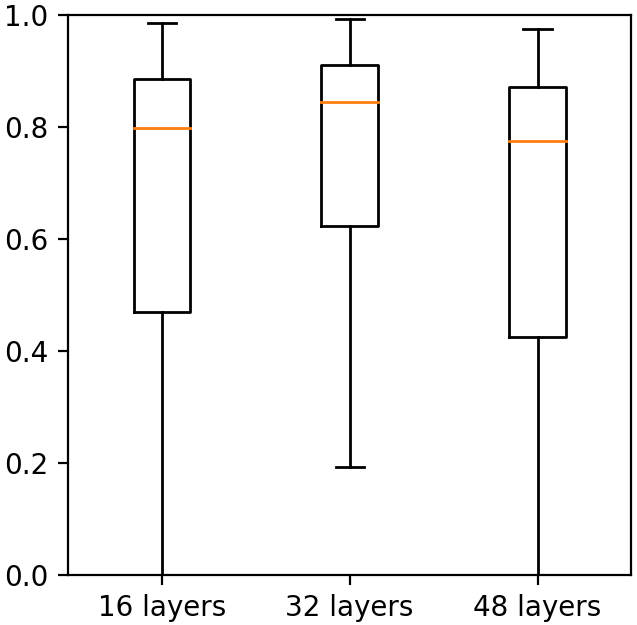}
            \caption{Pearson}
        \end{subfigure}
        % SSIM
        \begin{subfigure}[t]{0.24\columnwidth}
            \centering
            \includegraphics[width=\columnwidth]{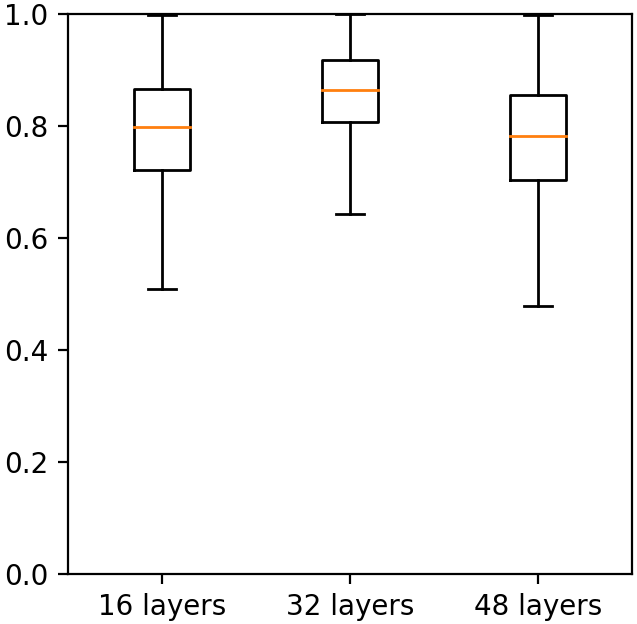}
            \caption{SSIM}
        \end{subfigure}
    \caption{Model performance of different number of layers. 
            It shows that the 32-layer one has lower MAE and RMSE, higher Pearson Correlation and SSIM. 
            It also implies the proper model size (number of parameters) in aspect of model capability and avoiding overfitting problems. }
    \label{fig:diffnsboxplot}
\end{figure}

        After training, we feed testing data for evaluations, and eventually pick the 32-layer one as our base, since it has the lowest MAE and RMSE and highest Corr. and SSIM, as shown in Fig. \ref{fig:diffnsboxplot}. 
        The configuration might vary with different data properties, but we offer a reference setting to the climate downscaling task.  

        % \subsection{Number of RABs}
        % As we finalize the total number of convolutional layers, here we are extracting intermediate feature maps from every \textit{\textbf{N}} layer into RAB module. 
        % Thus, this points to the number of RABs. 
        % The number of parameters of one RAB block is approximately 58,000, around 5\% of the mainstream convolutional layers' in our model. 
        % Since our input size is only 14$\times$9, we wonder if there are excessive or similar refined feature maps extracted that merely contribute to our final output. The results are shown in \cref{fig:diffnumRABboxplot}.

    \subsubsection{Image Upscaling Layers}
\begin{figure}[t!]
    \centering
        % MAE
        \begin{subfigure}[t]{0.24\columnwidth}
            \label{fig:diffupMAE}
            \includegraphics[width=\columnwidth]{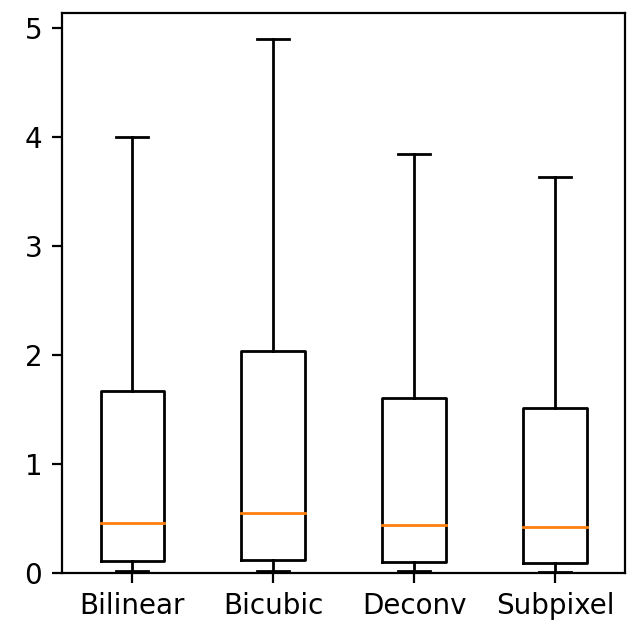}
            \caption{MAE}
        \end{subfigure}
        % RMSE
        \begin{subfigure}[t]{0.24\columnwidth}
            \centering
            \label{fig:diffupRMSE}
            \includegraphics[width=\columnwidth]{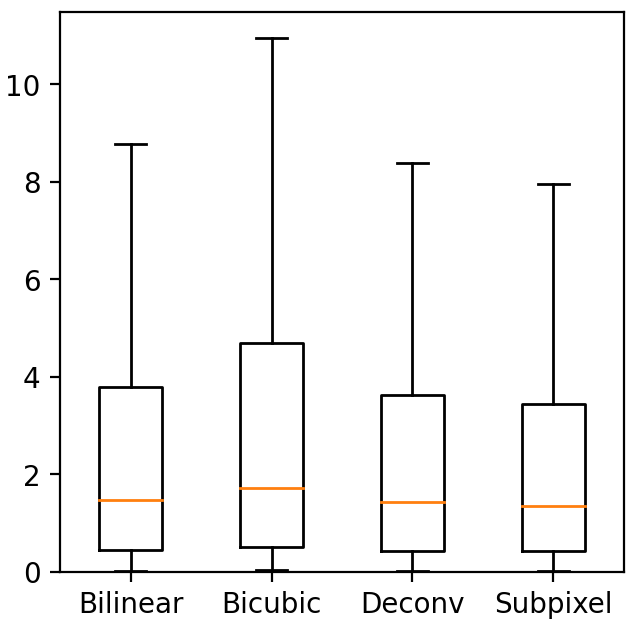}
            \caption{RMSE}
        \end{subfigure}
        % CORR
        \begin{subfigure}[t]{0.24\columnwidth}
            \centering
            \label{fig:diffupCorr}
            \includegraphics[width=\columnwidth]{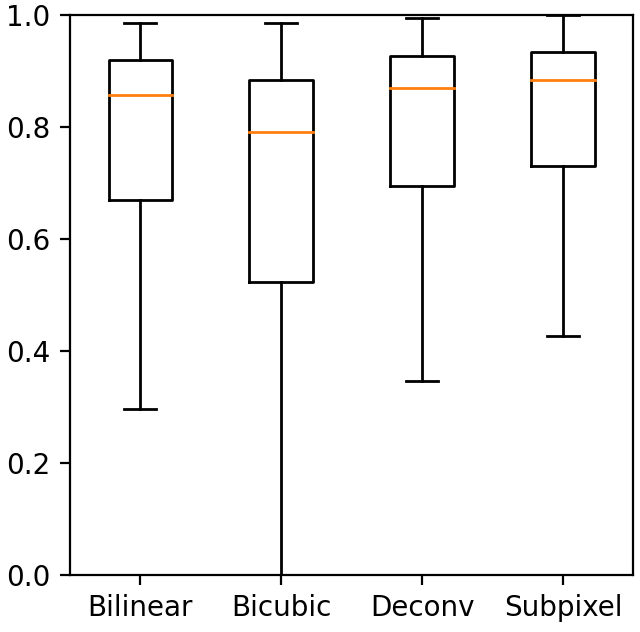}
            \caption{Pearson}
        \end{subfigure}
        % SSIM
        \begin{subfigure}[t]{0.24\columnwidth}
            \centering
            \label{fig:diffupSSIM}
            \includegraphics[width=\columnwidth]{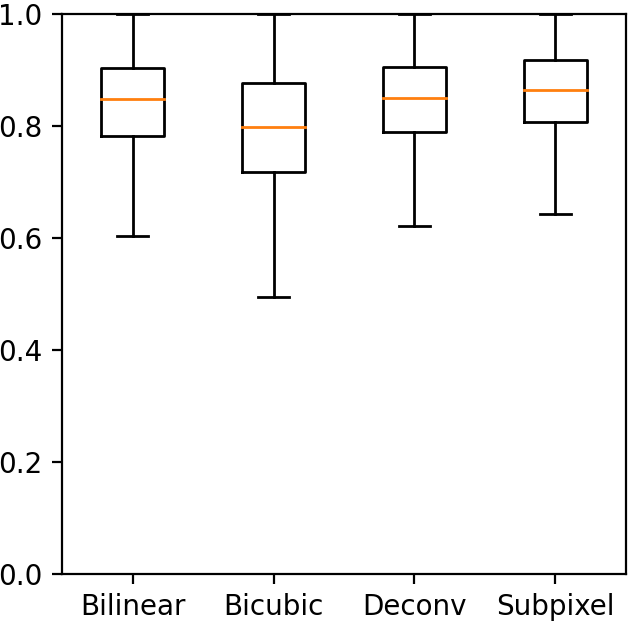}
            \caption{SSIM}
        \end{subfigure}
    \caption{Boxplots of different upscaling layers. 
            As mentioned in Sec. \ref{sec:method}, we show the performance of different types of image upscaling layer here. 
            It turns out to be that the pixel shuffling (subpixel) image upscaling method has the best prediction quality.}
    \label{fig:diffupboxplot}
\end{figure}

        As mentioned in Sec. \ref{sec:method}, different image upscaling layers would influence the quality of climate downscasling results. 
        Concerns are that bilinear and bicubic methods usually do not provide any ``new'' information to data, while deconvolutional layers and the pixel shuffling layer will encounter the checkerboard artifacts if the kernel size is not divisible by the number of strides (e.g., \cite{deker, ResLap}). 
        Although both bilinear and bicubic are non-learnable parameters in the model, they will influence other learnable ones. 
        For example, we were expecting our model to learn a low-resolution, but an intermediate output whose bias has been corrected right before the bilinear or bicubic layer. 
        Yet, the climate downscaling of heterogeneous data might be more complicated. 
        Hence, we verified four upscaling layers: bilinear, bicubic, deconvolutional layer, and pixel rearrangement to find the best-fitting one. 
        The results are shown in Fig. \ref{fig:diffupboxplot}. 
        The medians of bicubic one has the highest MAE and RMSE and the lowest Pearson Correlation and SSIM. 
        Besides, it has larger deviations, which shows the unstable predictions. 
        While, the pixel shuffling (pixel rearrangement) has the lowest MAE and RMSE, the highest Pearson Correlation and SSIM, and smaller deviation, showing relatively stable predictions. 
        Therefore, we adopt pixel shuffling layer as our image upscaling method. 
        
        \subsubsection{Topography Data}
\begin{figure}[t!]
    \centering
        % MAE
        \begin{subfigure}[t]{0.24\columnwidth}
            \centering
            \includegraphics[width=\columnwidth]{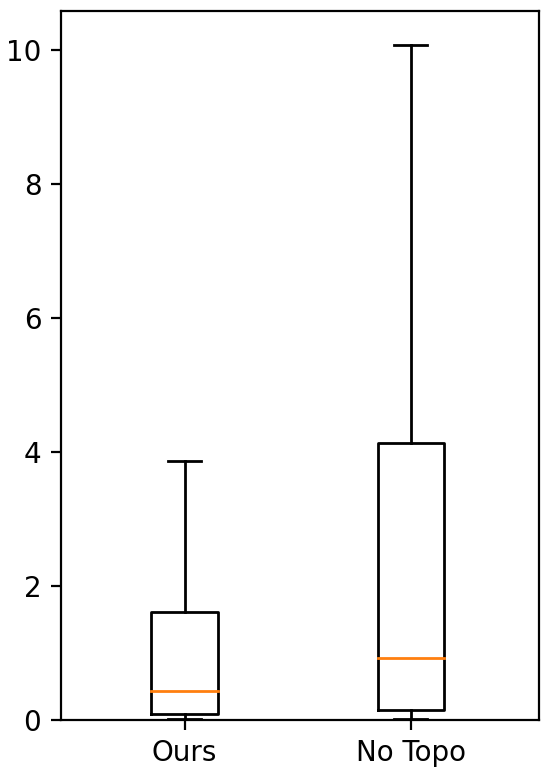}
            \caption{MAE}
        \end{subfigure}
        % \hfill
        % RMSE
        \begin{subfigure}[t]{0.24\columnwidth}
            \centering
            \includegraphics[width=\columnwidth]{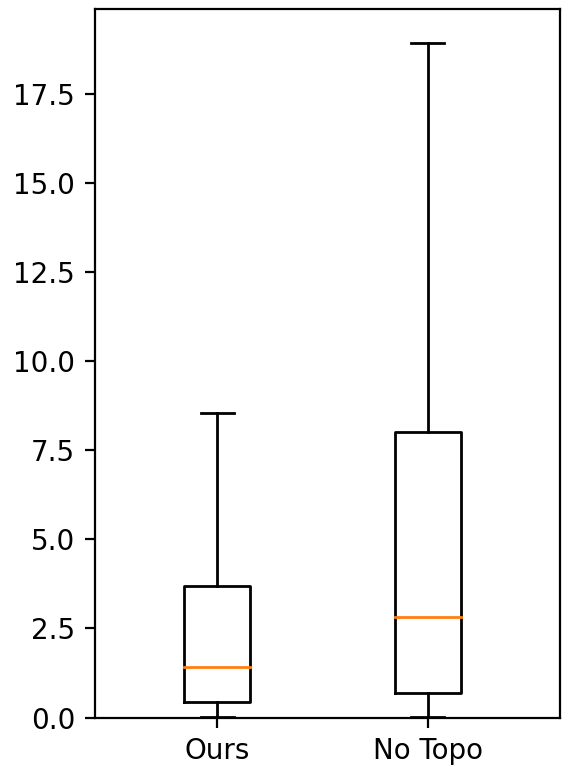}
            \caption{RMSE}
        \end{subfigure}
        % \vskip\baselineskip
        % CORR
        \begin{subfigure}[t]{0.24\columnwidth}
            \centering
            \includegraphics[width=\columnwidth]{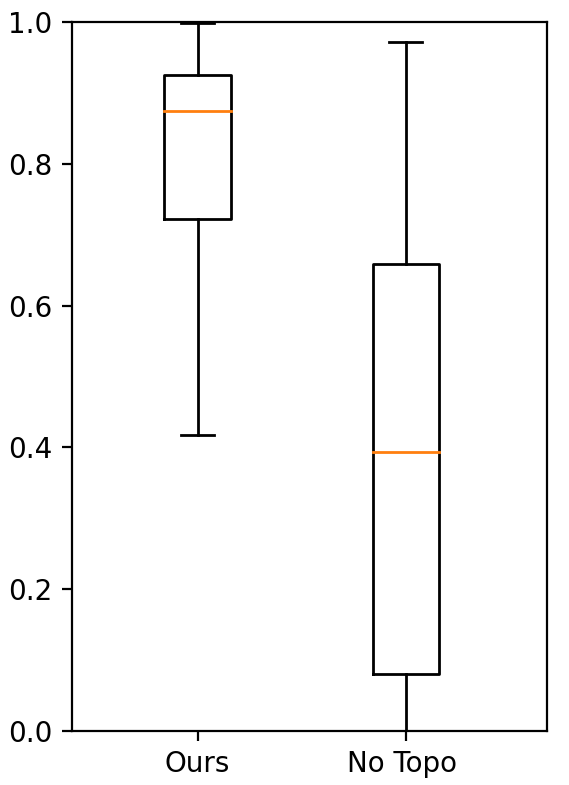}
            \caption{Pearson}
        \end{subfigure}
        % \hfill
        % SSIM
        \begin{subfigure}[t]{0.24\columnwidth}
            \centering
            \includegraphics[width=\columnwidth]{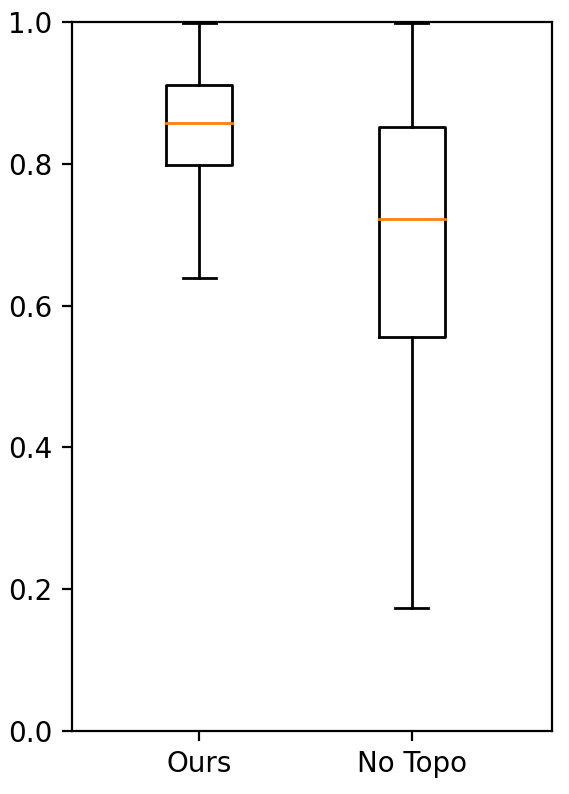}
            \caption{SSIM}
        \end{subfigure}
    \caption{Boxplots of whether to concatenate with topography. 
            It shows statistically significant difference in whether to adopt the topography data. 
            This implies there are strong influences in precipitation as mentioned in Sec. \ref{sec:intro}, and our model successfully capture such interaction between precipitation and topography.}
    \label{fig:topoboxplot}
\end{figure}

        As mentioned in Sec. \ref{sec:intro}, we show the enhancement of climate downscaling results by adding a concatenation layer with terrain data in Fig. \ref{fig:topoboxplot}.  
        It shows statistically significant difference in whether to concatenate with the topography data. 
        This implies that there are strong influences or correlations with precipitation, and that the neccessariness of topography data in the climate downscaling task. 
        Our model successfully capture such interaction between precipitation and topography, resulting in better prediction quality.

% example 3 - relatively evenly distributed precipitation
% MAEs of example 3
\newcolumntype{M}[1]{>{\centering\arraybackslash}p{#1}}
\begin{table*}
% \tiny
\resizebox{1.5\columnwidth}{!}{%
\centering
    \begin{tabular}{ccccccccc}
        \toprule
        \multirow[c]{2}[2]{*}{\shortstack{Scaling \\ Factor}} & \multirow[c]{2}[2]{*}{\shortstack{Downscaling \\ Method}} & \multicolumn4{c}{Metrics (Avg. / Med.)} & \multicolumn3{c}{Indicators (Avg. / Med.)} \\
        \cmidrule(lr){3-6} \cmidrule(lr){7-9}
         & & MAE & RMSE & Corr. & SSIM & POD & FAR & TS \\\midrule
         \multirow[c]{12}{*}{x2} 
            & \multirow[c]{2}{*}{QM} & 5.2078 & 9.0507 & 0.3532 & 0.4108 & 0.5919 & 0.4915 & 0.4129 \\
            & & 1.3397 & 3.8916 & 0.3364 & 0.3610 & 0.5000 & 0.4265 & 0.2937 \\
            \cmidrule(lr){2-9}
            & \multirow[c]{2}{*}{BCSD} & 5.2564 & 8.3366 & 0.3618 & 0.1043 & \textbf{1.0000} & 0.7170 & 0.3102 \\
                                     & & 1.9838 & 3.6999 & 0.3235 & 0.0609 & \textbf{1.0000} & 0.8306 & 0.1694 \\
            \cmidrule(lr){2-9}
            & \multirow[c]{2}{*}{DeepSD} & 4.8107 & 8.3481 & 0.3352 & 0.1310 & 0.9876 & 0.7037 & 0.3254 \\
                                       & & 1.1039 & 2.9429 & 0.3278 & 0.1179 & \textbf{1.0000} & 0.8049 & 0.1947 \\
            \cmidrule(lr){2-9}
            & \multirow[c]{2}{*}{FSRCNN-ESM} & 4.0904 & 7.1279 & 0.4601 & 0.2319 & 0.9952 & 0.7056 & 0.3233 \\
                                           & & 1.1011 & 2.7516 & 0.4616 & 0.2087 & \textbf{1.0000} & 0.8091 & 0.1907 \\
            \cmidrule(lr){2-9}
            & \multirow[c]{2}{*}{YNet} & 1.9920 & 4.1826 & 0.6978 & 0.6427 & 0.8491 & 0.4580 & 0.5708 \\
                                     & & 0.6409 & 2.0600 & 0.8179 & 0.6510 & 0.9143 & 0.4000 & 0.5345 \\
            \cmidrule(lr){2-9}
            & \multirow[c]{2}{*}{Ours} & \textbf{1.6884} & \textbf{3.5713} & \textbf{0.7526} & \textbf{0.6921} & 0.8428 & \textbf{0.4093} & \textbf{0.6052} \\
                                     & & \textbf{0.5429} & \textbf{1.7454} & \textbf{0.8703} & \textbf{0.7272} & 0.9091 & \textbf{0.3412} & \textbf{0.5801} \\
        \cmidrule(lr){1-9}
         \multirow[c]{12}{*}{x4} 
            & \multirow[c]{2}{*}{QM} & 5.4020 & 8.8380 & 0.3474 & 0.5739 & 0.5630 & 0.4801 & 0.3922 \\
                                   & & 1.3402 & 3.8548 & 0.3237 & 0.5516 & 0.4737 & 0.4375 & 0.2727 \\
            \cmidrule(lr){2-9}
            & \multirow[c]{2}{*}{BCSD} & 5.2238 & 8.2537 & 0.3539 & 0.2084 & \textbf{1.0000} & 0.7128 & 0.2983 \\
                                     & & 1.9871 & 3.6458 & 0.3198 & 0.1812 & \textbf{1.0000} & 0.8318 & 0.1682 \\
            \cmidrule(lr){2-9}
            & \multirow[c]{2}{*}{DeepSD} & 5.0753 & 8.9770 & 0.2822 & 0.2598 & 0.9724 & 0.7051 & 0.3089 \\
                                       & & 1.1408 & 2.9387 & 0.2664 & 0.2604 & \textbf{1.0000} & 0.8170 & 0.1821 \\
            \cmidrule(lr){2-9}
            & \multirow[c]{2}{*}{FSRCNN-ESM} & 3.9873 & 7.0270 & 0.4279 & 0.3455 & 0.9891 & 0.7010 & 0.3109 \\
                                           & & 1.1154 & 2.7751 & 0.4306 & 0.3296 & \textbf{1.0000} & 0.8068 & 0.1924 \\
            \cmidrule(lr){2-9}
            & \multirow[c]{2}{*}{YNet} & 1.9588 & 4.1685 & 0.6369 & 0.5634 & 0.7992 & 0.5278 & 0.4937 \\
                                     & & 0.6780 & 2.1344 & 0.7938 & 0.5647 & 0.9105 & 0.4930 & 0.4745 \\
            \cmidrule(lr){2-9}
            & \multirow[c]{2}{*}{Ours} & \textbf{1.6979} & \textbf{3.6271} & \textbf{0.6917} & \textbf{0.6625} & 0.8382 & \textbf{0.4598} & \textbf{0.5600} \\
                                     & & \textbf{0.5813} & \textbf{1.8998} & \textbf{0.8369} & \textbf{0.6624} & 0.9210 & \textbf{0.4098} & \textbf{0.5439} \\
        \cmidrule(lr){1-9}
        \multirow[c]{12}{*}{x5} 
            & \multirow[c]{2}{*}{QM} & 4.5550 & 7.9047 & 0.3766 & 0.6121 & 0.5138 & 0.4401 & 0.3636 \\
                                   & & 1.2571 & 3.5200 & 0.3660 & 0.5955 & 0.5655 & 0.3828 & 0.3474 \\
            \cmidrule(lr){2-9}
            & \multirow[c]{2}{*}{BCSD} & 4.4342 & 7.4289 & 0.4256 & 0.4067 & 0.6277 & 0.5356 & 0.3672 \\
                                     & & 1.8717 & 3.7533 & 0.4017 & 0.4059 & 0.7500 & 0.5301 & 0.3379 \\
            \cmidrule(lr){2-9}
            & \multirow[c]{2}{*}{DeepSD} & 5.0695 & 8.7594 & 0.2062 & 0.6116 & 0.6805 & 0.5952 & 0.3607 \\
                                       & & 0.8996 & 2.9104 & 0.1629 & 0.6445 & 0.7680 & 0.6262 & 0.3154 \\
            \cmidrule(lr){2-9}
            & \multirow[c]{2}{*}{FSRCNN-ESM} & 4.9565 & 8.6558 & 0.1801 & 0.6764 & 0.4899 & 0.3922 & 0.3672 \\
                                           & & 0.8655 & 2.9678 & 0.1211 & 0.7272 & 0.5377 & 0.3130 & 0.3487 \\
            \cmidrule(lr){2-9}
            & \multirow[c]{2}{*}{YNet} & 1.6101 & 3.4145 & 0.7112 & 0.7955 & \textbf{0.7208} & 0.3932 & 0.5405 \\
                                     & & 0.4616 & 1.4945 & 0.8573 & 0.8416 & \textbf{0.8728} & 0.3153 & 0.6083 \\
            \cmidrule(lr){2-9}
            & \multirow[c]{2}{*}{Ours} & \textbf{1.1469} & \textbf{2.9352} & \textbf{0.7391} & \textbf{0.8300} & 0.7149 & \textbf{0.3372} & \textbf{0.5635} \\
                                     & & \textbf{0.4172} & \textbf{1.3532} & \textbf{0.8832} & \textbf{0.8644} & 0.8658 & \textbf{0.2629} & \textbf{0.6384} \\
        \cmidrule(lr){1-9}
         \multirow[c]{12}{*}{x8} 
            & \multirow[c]{2}{*}{QM} & 5.0429 & 8.7130 & 0.3292 & 0.6855 & 0.4540 & 0.4855 & 0.3153 \\
                                   & & 1.3394 & 3.7770 & 0.2909 & 0.6969 & 0.4556 & 0.4557 & 0.2636 \\
            \cmidrule(lr){2-9}
            & \multirow[c]{2}{*}{BCSD} & 5.2495 & 8.2898 & 0.3407 & 0.2084 & \textbf{1.0000} & 0.7106 & 0.2894 \\
                                     & & 1.9730 & 3.6179 & 0.2934 & 0.1812 & \textbf{1.0000} & 0.8306 & 0.1694 \\
            \cmidrule(lr){2-9}
            & \multirow[c]{2}{*}{DeepSD} & 5.3036 & 9.3460 & 0.2598 & 0.4005 & 0.9632 & 0.6997 & 0.2994 \\
                                       & & 1.1991 & 3.0452 & 0.2338 & 0.3931 & 0.9981 & 0.8111 & 0.1883 \\
            \cmidrule(lr){2-9}
            & \multirow[c]{2}{*}{FSRCNN-ESM} & 4.5037 & 7.7278. & 0.3329 & 0.3609 & 0.9876 & 0.7050 & 0.2948 \\
                                           & & 1.2432 & 2.9157 & 0.3268 & 0.3577 & 0.9997 & 0.8218 & 0.1782 \\
            \cmidrule(lr){2-9}
            & \multirow[c]{2}{*}{YNet} & 1.9893 & 4.2271 & 0.6167 & 0.8230 & 0.6406 & 0.4227 & 0.4985 \\
                                     & & 0.5815 & 1.9557 & 0.7707 & 0.8501 & 0.8231 & 0.3193 & 0.5776 \\
            \cmidrule(lr){2-9}
            & \multirow[c]{2}{*}{Ours} & \textbf{1.7077} & \textbf{3.6412} & \textbf{0.6827} & \textbf{0.8647} & 0.6667 & \textbf{0.2955} & \textbf{0.5522} \\
                                     & & \textbf{0.4964} & \textbf{1.7104} & \textbf{0.8273} & \textbf{0.8849} & 0.8309 & \textbf{0.2183} & \textbf{0.6470} \\

        \bottomrule
    \end{tabular}
}
    \caption{Metrics and climate indicators of different climate downscaling methods with different scaling factors. 
            Metrics include MAE, RMSE, Pearson Correlation, and SSIM. 
            Climate indicators include POD, FAR, and TS. 
            The threshold of precipitation detection is set as 0.1 mm. 
            One has the best performance if it has lower MAE, RMSE and FAR as well as higher Corr., SSIM, POD and TS.}
    \label{tb:diffmodels}
\end{table*}
%%%%%%%%%%%%%%%%%%%%%%%%%%%%%%%%%%%%%%%%%%%%%%%%%%%%%%%%%%%%%%%%%
% Different Number of Layers

% Losses of Different Number of Layers

% % Different Number of RABs
% \begin{figure}[tb]
%     \centering
%         % MAE
%         \begin{subfigure}[t]{0.35\columnwidth}
%             \centering
%             \label{fig:diffnrabMAE}
%             \includegraphics[width=\columnwidth]{figures/Boxplot/DiffNumRABMAE.png}
%             \caption{MAE}
%         \end{subfigure}
%         \hfill
%         % RMSE
%         \begin{subfigure}[t]{0.35\columnwidth}
%             \centering
%             \label{fig:diffnrabRMSE}
%             \includegraphics[width=\columnwidth]{figures/Boxplot/DiffNumRABRMSE.png}
%             \caption{RMSE}
%         \end{subfigure}
%         \vskip\baselineskip
%         % CORR
%         \begin{subfigure}[t]{0.35\columnwidth}
%             \centering
%             \label{fig:diffnrabCorr}
%             \includegraphics[width=\columnwidth]{figures/Boxplot/DiffNumRABCORR.png}
%             \caption{Pearson Correlation}
%         \end{subfigure}
%         \hfill
%         % SSIM
%         \begin{subfigure}[t]{0.35\columnwidth}
%             \centering
%             \label{fig:diffnrabSSIM}
%             \includegraphics[width=\columnwidth]{figures/Boxplot/DiffNumRABSSIM.png}
%             \caption{SSIM}
%         \end{subfigure}
%     \caption{Boxplots of different number of RABs. 
%             It shows there is no significant difference among different numbers of RABs. 
%             Considering the number of hyperparameters and training time, we chose the 16-number-of-RABs one for the rest of our experiments.}
%     \label{fig:diffnumRABboxplot}
% \end{figure}

% Different Type of Upscaling Layer

% Ablation of topographical data

\section{Conclusion}
\label{sec:conclusion}
In this paper, we proposed a deep learning based climate downscaling model for the precipitation data in a strong climatic forcing, non-continental, and topographically drastic changing area. 
We have tackled with the data heterogeneity and severe bias problems between reanalysis data and observations. 
For our model training, we collected and picked ERA5 reanalysis data and TCCIP gridded observational data. 
Our model is mainly composed of a series of convolutional layers and utilizes several residual attention blocks, skip connections and concatenation of topography data to better catch the relations of regional climatic forcings to precipitation. 
Furthermore, we compared the predictions with two commonly-used statistical climate downscaling methods and other three machine learning based ones in metrics of MAE, RMSE, Corr. and SSIM, as well as the forecast indicators, POD, FAR and TS. 
And we have shown that our model has better performance among all the other methods.

% limitation and future work
Evidently, machine learning does bring skillful solutions to climate downscaling issues, but lack of physical, climatological explanations of how the climate patterns are determined. 
After all, the model is seen as a black box, and that is another interesting topic in interpretable or explainable AI field. 
For instance, by analyzing the intermediate layer outputs, we may use techniques like saliency maps to capture global or local explanations and help climatologists to fine-tune the simulation models. 
Combined with some visualization and interpretable techniques, it may help the climate science community to study uncertainties and climate patterns. 
For example, it would be more easily to manipulate climate analysis by developing a customized visualization tool which may also provide the public more comprehensive, straightforward, and understandable insights into our sophisticated climate systems. 
Another research direction could be an integrated model for the climate downscalings of simulation data like GCMs rather than reanalysis data to observations. 
After all, the cross-domain applications of machine learning are thriving nowadays. 
The potentiality is still awaiting the discovery and excavation.

% \input{paper/section7_Use Case.tex}
% \input{paper/section8_Discussion and conclusion.tex}
% \label{}

% \section*{Acknowledgement}
% We would like to thank the domain experts, Dr. Berlin Chen, YiCheng Wang, and HsinWei Wang, from National Taiwan Normal University to provide insight into NLP, transformer models, and feedback on our work.
\label{}
% aaa
% </geshi> 

% Numbered list
% Use the style of numbering in square brackets.
% If nothing is used, default style will be taken.
%\begin{enumerate}[a)]
%\item 
%\item 
%\item 
%\end{enumerate}  

% Unnumbered list
%\begin{itemize}
%\item 
%\item 
%\item 
%\end{itemize}  

% Description list
%\begin{description}
%\item[]
%\item[] 
%\item[] 
%\end{description}  

% % Figure
% \begin{figure}[<options>]
% 	\centering
% 		\includegraphics[<options>]{}
% 	  \caption{}\label{fig1}
% \end{figure}

% \begin{table}[<options>]
% \caption{}\label{tbl1}
% \begin{tabular*}{\tblwidth}{@{}LL@{}}
% \toprule
%   &  \\ % Table header row
% \midrule
%  & \\
%  & \\
%  & \\
%  & \\
% \bottomrule
% \end{tabular*}
% \end{table}

% Uncomment and use as the case may be
%\begin{theorem} 
%\end{theorem}

% Uncomment and use as the case may be
%\begin{lemma} 
%\end{lemma}

%% The Appendices part is started with the command \appendix;
%% appendix sections are then done as normal sections
%% \appendix

% \section{}\label{}

% To print the credit authorship contribution details
% \printcredits
\clearpage
%% Loading bibliography style file
% \bibliographystyle{abbrvnat}
\bibliographystyle{cas-model2-names}
% \bibliographystyle{chicago2}

% Loading bibliography database
\bibliography{citetion}

% % Biography
% \bio{}
% % Here goes the biography details.
% \endbio

% \bio{pic1}
% % Here goes the biography details.
% \endbio

\end{document}